\documentclass[a4paper,11pt]{article}

\usepackage{amsmath}
\usepackage{amssymb}  
\usepackage{latexsym} 
\usepackage{multirow}
\usepackage{tikz}

\usepackage{authblk}

\usepackage{fullpage}

\pgfdeclarelayer{edgelayer}
\pgfdeclarelayer{nodelayer}
\pgfsetlayers{edgelayer,nodelayer,main}

\tikzstyle{none}=[inner sep=0pt]
\tikzstyle{plain}=[inner sep=0pt]
\tikzstyle{every picture}=[baseline=(current bounding box).east,scale=0.5,node distance=5mm]

\newcommand{\ov}{\overrightarrow} 

\title{Reasoning about Meaning in Natural Language\\with Compact Closed Categories\\and Frobenius Algebras
\footnote{To appear in Chubb, J., Eskandarian, A. and Harizanov, V., editors, \textit{Logic and Algebraic Structures in Quantum Computing and Information} (2014), Cambridge University Press. Support by EPSRC grant EP/F042728/1 is acknowledged by the first two authors.
}}
\author{Dimitri Kartsaklis}
\author{Mehrnoosh Sadrzadeh}
\author{Stephen Pulman}
\author{Bob Coecke}

\affil{Department of Computer Science, University of Oxford\\
Wolfson Building, Parks Road, Oxford, OX1 3QD\\ 
{\tt firstname.lastname@cs.ox.ac.uk}
}   

\date{}

\begin{document}

\newcommand{\tikzfig}[1]{
\InputIfFileExists{#1.tikz}{}{\input{./#1.tikz}}
}

\newcommand{\ctikzfig}[1]{%
\begin{center}\rm
  
\InputIfFileExists{#1.tikz}{}{\input{./#1.tikz}}

\end{center}}

\maketitle 
    
\begin{abstract} 
\leftskip=0pt plus.5fil
\rightskip=0pt plus-.5fil
\parfillskip=0pt plus1fil
\noindent Compact closed categories have found applications in  modeling quantum information protocols by Abramsky-Coecke. They also provide semantics for  Lambek's pregroup algebras, applied to formalizing the grammatical structure of natural language, and are  implicit in a distributional model of word meaning based on vector spaces.  Specifically, in previous work Coecke-Clark-Sadrzadeh used the product category of pregroups with  vector spaces and provided a distributional model of meaning for sentences. We recast this theory in terms of strongly monoidal functors and advance it via Frobenius algebras over vector spaces. The former are used to formalize topological quantum field theories by Atiyah and  Baez-Dolan,  and the latter are used to model classical data in quantum protocols by Coecke-Pavlovic-Vicary. The Frobenius algebras enable us to work in a single space in which meanings of words, phrases, and sentences of any structure live. Hence we can compare meanings of different language constructs and enhance the applicability of the theory.  We report on experimental results on a number of language tasks  and verify the theoretical predictions. 
\end{abstract}

\section{Introduction}

Compact closed categories were first introduced by Kelly \cite{Kelly} in early 1970's. Some thirty years later  they found applications in quantum mechanics \cite{AbrCoe}, whereby the vector space foundations of quantum mechanics were recasted in a higher order language and quantum protocols such as \emph{teleportation} found succinct conceptual proofs. Compact closed categories  are complete with regard to a pictorial calculus ~\cite{Kelly,Selinger}; this calculus is used to depict and reason about information flows in  entangled quantum states modeled in tensor spaces, the phenomena that were considered to be mysteries of quantum mechanics and the Achilles heel of quantum logic \cite{vonNeumann}. The pictorial calculus revealed the multi-linear algebraic level needed for proving quantum information  protocols and simplified the reasoning thereof to a great extent, by hiding the underlying additive vector space structure. 

Most quantum protocols rely on classical, as well as quantum, data flow. In the work of  \cite{AbrCoe}, this classical data flow was modeled using bi-products defined over a compact closed category. However, the pictorial calculus could not extend well to bi-products, and their categorical axiomatization was not as clear as the built-in monoidal tensor of the category. Later, Frobenius algebras, originally used in group theory \cite{Frob} and later widely applied to other fields of mathematics and physics such as topological quantum field theory (TQFT) \cite{Atiyah,Kock,Baez}, proved useful. It turned out that the operations of such algebras on vector spaces with orthonormal basis correspond to a uniform copying and deleting of the basis, a property that only holds for, hence can be used to axiomatize, classical states \cite{CoeckeVic}. 

Compact closed categories have also found applications in two completely orthogonal areas of computational linguistics: formalizing grammar and reasoning about lexical meanings of words. The former application is through Lambek's pregroup grammars \cite{Lambek}, which  are compact closed categories \cite{PrellerLambek} and have been applied to formalizing grammars of a wide range of natural languages, for instance see \cite{CasadioLambek}. The other application domain, referred to as \emph{distributional} models of meaning,  formalizes meanings of words regardless of their grammatical roles and via the context of their occurrence \cite{Firth}.  These models consist of vector spaces whose basis are sets of context words and whose vectors  represent meanings of target words. Distributional models have been widely studied and successfully applied to a variety of language tasks \cite{Schutze,Landauer,Manning} and in particular to automatic word-synonymy detection \cite{Curran}. 

Whereas the type-logical approaches to language do  not provide a convincing model of word meaning, the distributional models do not scale to meanings of phrases and sentences. The long standing challenge of combining these two models  was addressed in previous work \cite{ClarkPulman,Coeckeetal,PrellerSadr}. The solution was based on a cartesian product of the pregroup category and the category of finite dimensional vector spaces. The  theoretical predictions of the model were made concrete in \cite{Grefenetal}, then implemented and verified in \cite{GrefenSadr1}. In this article, we first recast the theoretical setting of \cite{Coeckeetal} using a  succinct functorial passage from a free pregroup of basic types to the category of finite dimensional vector spaces.  Then,  we further advance the theory and show how Frobenius algebras over vector spaces provide solutions for the problem of the concrete construction of linear maps for predicative words with complex types.  As a result,  we are able to compare meanings of phrases and sentences with different structures, and moreover compare these with lexical vectors of words. This enhances the domain of application of our model: we  show how  the theoretical predictions of the model, and in particular the Frobenius algebraic constructions, can be empirically verified by performing three experiments: the disambiguation  task of \cite{GrefenSadr1}, comparing meanings of transitive and intransitive sentences, and a new term/definition classification task. 

\section{Recalling Some Categorical Definitions}
\label{sec:abs}

We start by recalling some definitions. A \emph{monoidal category} \cite{Kelly} is a category ${\cal C}$ with a monoidal tensor $\otimes$, which is associative. That  is,  for all objects $A,B, C \in {\cal C}$, we have that $A \otimes (B \otimes C) \cong (A \otimes B) \otimes C$. Moreover there exists an object $I \in {\cal C}$, which serves as the unit of the tensor, that is, $A \otimes I \cong A \cong I \otimes A$. These isomorphisms need to satisfy the usual coherence conditions. 

A monoidal category is called \emph{symmetric} whenever we have $A \otimes B \cong B \otimes A$, again satisfying the standard  conditions. Furthermore, a monoidal category is \emph{compact closed} whenever any object $A \in {\cal C}$ has a left $A^l$ and a right adjoint $A^r$, that is, the following morphisms exist:
\begin{align*}
\epsilon^r_A \colon A \otimes A^r  \to I \qquad&\qquad  \eta^r_A \colon I \to A^r \otimes A\\
\epsilon^l_A \colon A^l \otimes A \to I \qquad&\qquad \eta^l_A \colon I \to A \otimes A^l
\end{align*}
and they satisfy the following \emph{yanking} conditions:
\begin{align*}
(1_A \otimes \epsilon_A^l) \circ (\eta^l_A \otimes 1_A) = 1_A \qquad & \qquad 
(\epsilon^r_A \otimes 1_A) \circ (1_A \otimes \eta^r_A) = 1_A\\
(\epsilon^l_A \otimes 1_{A^l}) \circ (1_{A^l} \otimes \eta^l_A) = 1_{A^l} \qquad &\qquad 
(1_{A^r} \otimes \epsilon^r_{A}) \circ (\eta^r_{A} \otimes 1_{A^r}) = 1_{A^r}
\end{align*}
In a \emph{symmetric compact closed} category, the left and right adjoints collapse into one, that is  we have $ A^* :=  A^l =  A^r$ and the above four equalities collapse to the following two:
\[
(\epsilon_A \otimes 1_A) \circ (1_A \otimes \eta_A)= 1_A \qquad \qquad 
(1_{A^*} \otimes \epsilon_A) \circ (\eta_A \otimes 1_{A^*})   = 1_{A^*}
\]

A functor ${\cal F}$ from a monoidal category ${\cal C}$ to a monoidal category ${\cal D}$ is a \emph{monoidal functor}  \cite{KockStrong},  whenever ${\cal F}$ is a functor and moreover there exists a morphism $I \to {\cal F}(I)$ and the following  is a natural transformation:
\[
{\cal F} (A) \otimes {\cal F}(B) \to {\cal F}(A \otimes B)
\]
 satisfying the corresponding coherence conditions. A monoidal functor is \emph{strongly monoidal} \cite{KockStrong}, whenever the above morphism and natural transformation are invertible.    

A strongly monoidal functor on two compact closed categories ${\cal C}$ and ${\cal D}$ preserves the compact structure, that is ${\cal F}(A^l) = {\cal F}(A)^l$ and ${\cal F}(A^r) = {\cal F}(A)^r$. To see this, consider the case of the left adjoint, for which we have the following two compositions of morphisms:
\begin{align*}
{\cal F}(A^l) \otimes {\cal F}(A) \  &\to {\cal F}(A^l \otimes A) \to {\cal F}(I) \to I\\
I \  &\to {\cal F}(I) \to {\cal F}(A \otimes A^l) \to {\cal F}(A) \otimes {\cal F}(A^l)
\end{align*}
From these, and since adjoints are unique, it follows that ${\cal F}(A^l)$ must be left adjoint to ${\cal F}(A)$. The case for the right adjoint is similar. 

%

An example of a compact closed category is a Lambek pregroup \cite{Lambek}, denoted by $(P, \leq, 1, \cdot, (-)^l, (-)^r)$; we refer to this category by {\bf Preg}. This is a partially ordered  monoid where each element of the partial order has a left and a right adjoint, that is we have the following inequalities, which are the partial order versions of the yanking conditions of a compact closed category:
\[
p\cdot p^r \leq 1 \leq p^r \cdot p \hspace{2cm} 
p^l \cdot p \leq 1 \leq p \cdot p^l
\]

An example of a pregroup is the set of all unbounded monotone functions on integers, with function composition as the monoidal tensor and the identity function as its unit. The left and right adjoints are defined using the standard definition of adjoints and in terms of the min and max operations on the integers as follows, for $f \in \mathbb{Z}^{\mathbb{Z}}$ and $m,n \in \mathbb{Z}$:
\[
f^r(n) = \bigvee \{m \in \mathbb{Z} \mid f(m) \leq n\} \hspace{2cm}
f^l(n) = \bigwedge \{m \in \mathbb{Z} \mid n \leq f(m)\}
\]

An example of a symmetric compact closed category is the category of finite dimensional vector spaces and linear maps over a field (which for our purposes we take to be the set of real numbers $\mathbb{R}$); we refer to this category by {\bf FVect}. The monoidal tensor is the tensor product of vector spaces whose unit is the field. The adjoint of each vector space is its dual, which, by fixing a basis  $\{r_i\}_i$, becomes isomorphic to the vector space itself, that is we have $A^* \cong A$ (note that this isomorphism is not natural). The $\epsilon$ and $\eta$ maps, given by the inner product and maximally entangled states or \emph{Bell pairs}, are defined as follows:
\begin{align*}
\epsilon_A\colon A \otimes A \to \mathbb{R} \qquad & \quad \mbox{given by} \quad \sum_{ij}\  c_{ij} \ r_i \otimes r_j \ \mapsto \ \sum_{ij} c_{ij} \langle r_i \mid r_j \rangle\\
\eta_A \colon \mathbb{R} \to A \otimes A \qquad & \quad \mbox{given by} \quad  1 \ \mapsto \ \sum_i r_i \otimes r_i
\end{align*}

An example of a monoidal functor is Atiyah's definition of a topological quantum field theory (TQFT). This is a representation of category of manifolds and cobordisms {\bf Cob} (representing, respectively possible choices of space and spacetime) over the category of finite dimensional vector spaces {\bf FVect}. This representation is formalized using a strongly monoidal functor from {\bf Cob} to {\bf FVect} by Baez and Dylon \cite{Baez}, and  assigns a vector space of states to each manifold and a linear operation to each cobordism. 

\section{Category Theory in Linguistics}
\label{sec:conc}

We briefly review two orthogonal models of meaning in computational linguistics: pregroup grammars and \emph{distributional} models of meaning, and we show how one can interpret the former  in the latter  using a strongly monoidal  functor. 

\subsection{Type-Logical Pregroup Grammars}

Consider the simple grammar generated by the following set of rules:

\vspace{-0.4cm}
\begin{center}
\begin{minipage}{3cm}
\begin{align*}
S &\to Np \ Vp  \\
Vp &\to tV \ Np \mid N\\
Np &\to Adj \ Np \mid N
\end{align*}
\end{minipage}
\hspace{2cm}
\begin{minipage}{3cm}
\begin{align*}
itV &\to \text{smile}\\
tV & \to \text{build}\\
Adj & \to \text{strong}\\
N & \to \text{man}, \text{woman},  \text{house}
\end{align*}
\end{minipage}
\end{center}

\noindent The above rules are referred to as \emph{generative rules}. The rules on the left describe the formation of a grammatical sentence $S$ in terms of other non-terminals.  According to these rules, a sentence is a noun phrase $Np$ followed by a verb phrase $Vp$, where a verb phrase itself is a transitive verb $tV$ followed either by a $Np$ or a noun $N$, and a noun phrase is an adjective $Adj$ followed either by a $Np$ or a noun $N$. The rules on the right instantiate all but one ($S$) of the non-terminals to terminals. According to these, `smile' is an intransitive verb, `build' is a transitive verb, `strong' is an adjective, and `man', `woman', and `house' are nouns. We treat these words as lemmas and take freedom in conjugating them in our example sentences. 

In a predicative approach, the non-terminals of the above grammar (except for $S$) are interpreted as unary or binary predicates to produce meaning for phrases and sentences. There are various options when  interpreting these non-terminals: for instance, according to the first rule, we can either interpret a verb phrase as a binary predicate that inputs a noun phrase and outputs a sentence, or we can interpret a noun phrase as a binary predicate that inputs a verb phrase and outputs a sentence. We adhere to the more popular (among computational linguistics) verb-centric view and follow the former option. The types of the resulting predicates, obtained by recursively unfolding the rules, form an algebra of types, referred to as a \emph{type-logical grammar}. 

A pregroup type-logical grammar, or  a pregroup grammar for short, is the pregroup freely generated over  a set of basic types which, for the purpose of this paper, we take to be  $\{n,s\}$. We refer to this free pregroup grammar by {\bf Preg$_F$}. Here,  $n$ is the type representing a noun phrase and $s$ is the type representing  a sentence. The complex types of this pregroup represent  the predicates. For instance, $n^r \cdot s$ is the type of an intransitive verb, interpreted as a unary predicate that inputs a noun phrase and outputs a sentence. Explicit in this type is also the fact that the intransitive verb has to be on the right hand side of the noun phrase. This fact is succinctly expressed by  the adjoint $r$ of the type $n$. Similarly, $n^r\cdot s\cdot n^l$ is  the type of a transitive verb, which is a binary predicate that inputs two noun phrases, has to be to the right of one and to the left of the other, and outputs a sentence. Finally, $n\cdot n^l$ is the type of an adjective in attributive position, a unary predicate that inputs a noun phrase and outputs another noun phrase; furthermore, it has to be to the left of its input noun phrase. These types are then assigned to the vocabulary of a language, that is to the non-terminals of the generative rules, via a relation referred to as a  \emph{type dictionary}. Our example type dictionary is as follows:
\begin{center}
\begin{tabular}{c|c|c|c|c|c}
man\  & woman \ & \ houses \ & \ strong \ & \ smiled \ & \ built\\ \hline
&&&&&\\
$n$& $n$ & $n$ & $n\cdot n^l$ & $n^r \cdot s$ & $n^r\cdot s \cdot n^l$\\
&&&&&\\
\end{tabular}
\end{center}

Every sequence of words $w_1 w_2 \cdots w_n$ from the vocabulary has an associated \emph{type reduction}, to which we refer to by $\alpha_{w_1 w_1 \cdots w_n}$. This type reduction represents the grammatical structure of the sequence. In a pregroup grammar, a type reduction is the result of applying the partial order, monoid, and adjunction axioms to the multiplication of the types of the words of the sequence. For example,  the type reduction $\alpha_{\mbox{\small strong house}}$ associated to the sequence `strong house' is computed by multiplying the types of `strong' and `house', that is $n \cdot n^l \cdot n$, then applying to it the adjunction and monoid axioms, hence obtaining $n \cdot n^l \cdot n \leq  n$. Similarly, the type reduction of the sentence `strong man built houses' is as follows:
\[
\alpha_{\mbox{\small strong man built houses}} \quad \colon \quad n \cdot n^l \cdot n \cdot n^r \cdot s \cdot n^l \cdot n \ \leq  \ n \cdot n^r \cdot s \ \leq \  s 
\]
In categorical terms, the type reduction is a morphism of the category {\bf Preg$_F$}, denoted by tensors and compositions of  the $\epsilon$ and identity maps. For instance, the morphisms corresponding to the above adjective-noun phrase and sentence are as follows:
\begin{center}
\begin{tabular}{c|c}
strong man \ & \ strong man built houses \\
\hline
&\\
 $1_n \otimes \epsilon^l_n$ \ & \  $(\epsilon^r_n \otimes 1_s) \circ (1_n \otimes \epsilon^l_n \otimes 1_{n^r\cdot s} \otimes \epsilon^l_n)$\\
 &\\
\end{tabular}
\end{center}

\medskip
The generative rules formalize the grammar of a natural language and  their consequent type-logical grammars provide a predicative  interpretation for the words with  complex types. However,  all the words with the same type have the same interpretation, and even worse,  words with basic types are only interpreted as atomic symbols. In the next section, we will see how \emph{distributional} models of meaning address this problem. 

\subsection{Distributional Models of Word Meaning}

Meanings of some words can be determined by their denotations. For instance,  meaning of the word `house' can be the set of all  houses or their images; and the answer to the question `what is a house?' can be provided by pointing to a house. Matters get  complicated when it comes to words with complex types such as adjectives and verbs. It is not so clear what is  the denotation of the adjective `strong' or the verb `build'. The problem is resolved by adhering to a meaning-as-use model of meaning,  whereby one can assign meaning to all words, regardless of their grammatical type, according to the context in which they often appear. This context-based approach to meaning is the main idea behind the \emph{distributional} models of meaning. 

First formalized by Firth in 1957  \cite{Firth} and about half a century later implemented and applied to word sense disambiguation by Sch\"utze \cite{Schutze}, distributional models of meaning interpret words as vectors in a highly dimensional (but finite) vector space with a fixed orthonormal basis over real numbers. A basis for this vector space is a set of \emph{target} words, which in principle can be the set of all lemmatized words of a corpus of documents or a dictionary. In practice, the basis vectors are often restricted to the few thousands most occurring words of the corpus, or a set of specialized words depending on the application domain, e.g. a music dictionary. Alternatively, they can be \emph{topics} obtained from a dimensionality reduction algorithm such as \emph{single value decomposition} (SVD). We refer to such a vector space with an orthonormal basis $\{w_i\}_i$, no matter how it is built, as our basic distributional vector space $W$; and to {\bf FVect} restricted to tensor powers of $W$ as {\bf FVect$_W$}. 

In this model, to each word is associated a vector, which serves as the meaning of the word. The weights of this vector are  obtained by counting how many times the word has appeared `close' to a basis word, where `close' is a window of $n$ (usually equal to 5) words. This number is usually normalized, often in the form of {\sc Tf-Idf} values which show how important is a specific basis word by taking into account not only the number of times it has occurred to the document, but also the number of documents in which it appears in the corpus.

  
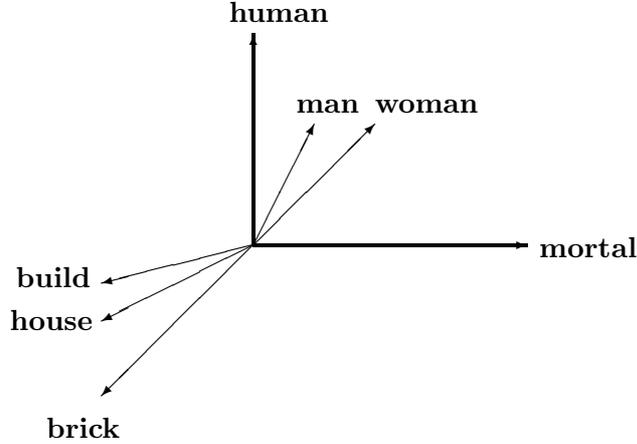
\begin{figure}[h]
\begin{center}
\begin{minipage}{10cm}
\begin{center}
\setlength{\unitlength}{0.8mm}
\begin{picture}(60, 70)
  \linethickness{0.3mm}
  \put(26,57){\text{\bf human}}
  \put(30, 20){\vector(1, 0){45}}
  \put(77,18){\text{\bf mortal}}
  \put(30, 20){\vector(0, 1){35}}
  \put(30, 20){\vector(-1, -1){25}}
  \put(-4,-12){\text{\bf brick}}
   
  \linethickness{0.3mm}
   
  \put(30, 20){\begin{color}{black}\vector(1, 1){20}\end{color}}
  \put(37,42){\text{\bf \begin{color}{black}man\end{color}}}
  
  \put(30, 20){\begin{color}{black}\vector(1, 2){10}\end{color}}
  \put(50,42){\text{\bf \begin{color}{black}woman\end{color}}}
  
  \put(30, 20){\begin{color}{black}\vector(-2, -1){25}\end{color}}
  \put(-9,13){\text{\bf \begin{color}{black}build\end{color}}}
  
  \put(30, 20){\begin{color}{black}\vector(-4, -1){25}\end{color}}
  \put(-10,6){\text{\bf \begin{color}{black}house\end{color}}}
 \end{picture}
\end{center}

\vspace{0.9cm}
\end{minipage}
\end{center}
\caption{A Toy  Distributional Model  of Meaning}
\label{fig:VectMod}
\end{figure}

As an example, consider the  toy vector space of Figure \ref{fig:VectMod}. The set $\{\text{human}, \text{mortal}, \text{brick}\}$ is the basis of this vector space and the words `man', `woman', `house' and `build' each have a vector assigned to them. The  words that have often appeared in the same context  have a smaller angle between their vectors. For instance, `house' and `build' often appear close to `brick', whereas `man' and `woman' often appear close to `mortal' and `human'. The cosine of the angle between the word vectors has proved to be a good measure in predicting  synonymy of words \cite{Curran}. Despite these good predictions, the distributional models of meaning cannot serve as the definite models of natural language, as there is more to a language than the contexts of its words and these models on their own do not scale up to the interpretations of phrases and sentences. In the next section, we will see how a combination of type-logical and distributional models  overcome both of their corresponding shortcomings. 

\subsection{Quantizing the Grammar}

We provide a mapping of the free pregroup grammar {\bf Preg$_F$} to {\bf FVect$_W$} via a strongly monoidal functor ${\cal F}$. This functor  assigns the basic vector space $W$ to both of the basic types, that is, we have:
\[
{\cal F}(n) = {\cal F}(s) = W
\]
By functoriality, the partial orders between the basic types (for example those presented in \cite{Lambek}) are mapped to linear maps from $W$ to $W$. The adjoints of basic types are also mapped to $W$, since for $x \in \{n,s\}$ we have the following, motivated by the above mentioned fact that $W^* \cong W$:
\[
{\cal F}(x^l) = {\cal F}(x^r) =  {\cal F}(x)
\]
Since $W^{**} \cong W^* \cong W$,  the iterated adjoint types are also mapped to $W$:
\[
{\cal F}(x^{ll}) = {\cal F}(x^{rr}) = {\cal F}(x)
\]
The complex types are mapped to tensor products of vector spaces, that is:
\[
{\cal F}(n \cdot n^l) = {\cal F}(n^r \cdot s) = W \otimes W \qquad
{\cal F}(n^r \cdot s \cdot n^l) = W \otimes W \otimes W
\]
Similarly, the type reductions are mapped to the compositions of tensor products of identity and  $\epsilon$ maps of {\bf FVect$_W$}, for instance  the  type reduction of a transitive sentence is mapped as follows:
\[
{\cal F}(\alpha_{\mbox{\small sbj verb obj}}) = {\cal F}(\epsilon^r_n \otimes 1_s \otimes \epsilon^l_n) = \epsilon_W \otimes 1_W \otimes \epsilon_W   \colon W \otimes (W \otimes W \otimes W) \otimes W \longrightarrow W
\]

Now we can use the definition of \cite{Coeckeetal} to provide a meaning for phrases and sentences of our grammar. The meaning of a sequence of words $w_1 w_2 \cdots w_n$ with type reduction $\alpha_{w_1 w_2 \cdots w_n}$ is:
\[
\text{\bf Definition(*)} \quad {\cal F}(\alpha_{w_1 w_2 \cdots w_n}) \left(\ov{w_1} \otimes \ov{w_2} \otimes \cdots \otimes \ov{w_n}\right)
\]
As an example, take:
\[
\ov{men} = \sum_i c_i^{men} \ov{w_i}\qquad
\ov{houses} = \sum_k c_k^{houses} \ov{w_k} \qquad
\overline{built} = \sum_{ijk}c^{built}_{ijk}(\ov{w_i} \otimes \ov{w_j} \otimes \ov{w_k})
\]
Substituting these in Definition(*), we obtain the following for the meaning of the sentence `men built houses':
\begin{align*}
{\cal F}\left(\epsilon^r_n \otimes 1_s \otimes \epsilon^l_n \right) \left(\ov{men} \otimes \overline{built} \otimes \ov{houses}\right) &= \\ 
\left(\epsilon_W \otimes 1_W \otimes \epsilon_W \right) \left(\ov{men} \otimes \overline{built} \otimes \ov{houses}\right) &=\\
\sum\limits_{ijk}  c^{built}_{ijk} \langle \ov{men}|\ov{w_i} \rangle \langle \ov{w_k}|\ov{houses}\rangle \ov{w_j}
\end{align*}

This definition ensures that the interpretations of noun phrases and sentences of any grammatical structure, for instance intransitive or transitive, will be a vector in $W$, hence we can measure the cosine of the angle between them and compute their synonymy. In order to determine that this measure of synonymy provides good predictions, we need to run some experiments. However, whereas we know very well how to build vectors in $W$ for words with basic types such as `man' and `house', our method further requires interpretations of words with complex types to be in tensor spaces, and there is no known standard procedure to construct these. In the next section we show how the notion of a Frobenius algebra over a vector space can be of use in addressing this matter. 

\section{Frobenius Algebras}
\label{sec:frob}

Frobenius algebras were originally introduced in 1903 by F.~G.~Frobenius in the context of proving representation theorems for group theory \cite{Frob}. Since then,  they have found applications in other fields of mathematics and physics, e.g. in topological quantum field theories \cite{Kock} and in categorical quantum mechanics \cite{CoeckeVic}. The  general categorical definitions recalled below are due to Carboni and Walters \cite{CarboniWal}. Their concrete instantiations to algebras over vector spaces were developed in \cite{CoeckeVic}. 

A Frobenius algebra over a symmetric monoidal category $({\cal C}, \otimes, I)$ is a tuple $(F, \sigma, \iota, \mu, \zeta)$, where for an $F$ object of ${\cal C}$ the triple $(F, \sigma, \iota)$  is an associative coalgebra, that is, the following are morphisms of ${\cal C}$, satisfying the coalgebraic associativity and unit conditions:

\vspace{-0.4cm}
{
\begin{align*}
\sigma \colon F \to F \otimes F&\qquad& \iota \colon F \to I
\end{align*}}

\noindent The triple $(F, \mu, \zeta)$ is an associative algebra, that is, the following are morphisms of ${\cal C}$, satisfying the algebraic associativity and unit conditions:

\vspace{-0.4cm}
{
\begin{align*}
\mu \colon  F \otimes F \to F  &\qquad& \zeta \colon I \to F 
\end{align*}}

\vspace{-0.4cm}
\noindent
Moreover, the above $\sigma$ and $\mu$ morphisms satisfy the following \emph{Frobenius} condition:

\vspace{-0.4cm}
\begin{align*}
(\mu \otimes 1_F) \circ (1_F \otimes \sigma) \ = \  \sigma \circ \mu  \ = \  (1_F \otimes \mu) \circ (\sigma \otimes 1_F) 
\end{align*}

A Frobenius Algebra is \emph{commutative} if it satisfies the following two conditions for $\delta: X \otimes Y \to Y \otimes X$, the symmetry morphism of $({\cal C}, \otimes, I)$:
\[
\delta \circ \sigma = \sigma \hspace{2cm} \mu \circ \delta = \mu
\]

\noindent Finally, a Frobenius Algebra is \emph{isometric} or \emph{special} if it satisfies the following condition:
\[
\mu \circ \sigma = Id
\]

\noindent In the category \textbf{FVect}, any vector space $V$ with a fixed basis $\{\ov{v_i}\}_i$ has a commutative special Frobenius algebra over it, explicitly given as follows:

\vspace{-0.4cm}
\begin{align*}
\sigma :: \ov{v_i} \mapsto \ov{v_i} \otimes \ov{v_i} &\qquad& \iota:: \ov{v_i} \mapsto 1\\
\mu:: \ov{v_i} \otimes \ov{v_i} \mapsto \ov{v_i} &\qquad& \zeta:: 1 \mapsto \ov{v_i}  
\end{align*}

\noindent In a Frobenius algebra over an orthonormal vector space, the coalgebra and algebra operations relate to each other via the equation $\sigma^{\dagger} = \mu$, where $\dagger$ is the adjoint, equal to the transpose for vector spaces over reals. 

In such Frobenius algebras, the operation $\sigma$  corresponds to copying and its unit $\iota$ corresponds to  deleting of the vectors. They enable one to faithfully encode vectors of $W$ into spaces with higher tensor ranks, such as $W \otimes W, W \otimes W \otimes W, \cdots$.  In linear algebraic terms, for $v \in W$, we have that $\sigma({v})$ is a diagonal matrix whose diagonal elements are weights of ${v}$.  The operation $\mu$ is referred to as uncopying; it loses some information when encoding a higher rank tensor into a lower rank space. In linear algebraic terms, for ${z} \in W\otimes W$, we have that $\mu({z})$ is a vector consisting only of the diagonal elements of ${z}$, hence losing the information encoded in the non-diagonal part. 

\section{Pictorial Calculi} 

The  framework of compact closed categories comes with a complete diagrammatic calculus that allows convenient graphical representations of the derivations. We briefly introduce the fragment of this calculus that we are going to use in this paper. The objects of this fragment  are the \textit{tensors} of multi-linear algebra; that is, a vector is a rank-1 tensor, a matrix is a rank-2 tensor, and a 3d-array is a rank-3 tensor. Each tensor is represented by a triangle, whose rank can be determined by its wires. Words are represented by tensors that correspond to their meaning: subjects and objects are rank-1 tensors (vectors), adjectives and intransitive verbs are rank-2 tensors, and transitive verbs are rank-3 tensors. The $\epsilon$ maps are depicted as cups, whereas the identity morphism is a vertical straight line. The tensor products of vectors are represented by juxtaposing their corresponding triangles. For example, the meaning of a transitive sentence, following Definition(*), is depicted as follows:

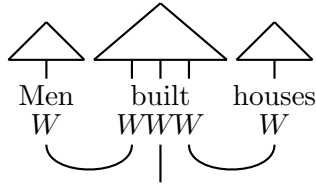
\begin{center}
  
\begin{tikzpicture}
	\begin{pgfonlayer}{nodelayer}
		\node [style=none] (0) at (3.75, 3.5) {};
		\node [style=none] (1) at (0.75, 3) {};
		\node [style=none] (2) at (6.75, 3) {};
		\node [style=none] (3) at (-0.25, 2) {};
		\node [style=none] (4) at (0.75, 2) {};
		\node [style=none] (5) at (1.75, 2) {};
		\node [style=none] (6) at (2, 2) {};
		\node [style=none] (7) at (3.000001, 2) {};
		\node [style=none] (8) at (3.75, 2) {};
		\node [style=none] (9) at (4.5, 2) {};
		\node [style=none] (10) at (5.5, 2) {};
		\node [style=none] (11) at (5.75, 2) {};
		\node [style=none] (12) at (6.75, 2) {};
		\node [style=none] (13) at (7.75, 2) {};
		\node [style=none] (14) at (0.75, 1.5) {};
		\node [style=none] (15) at (3, 1.5) {};
		\node [style=none] (16) at (3.75, 1.5) {};
		\node [style=none] (17) at (4.5, 1.5) {};
		\node [style=none] (18) at (6.75, 1.5) {};
		\node [style=none, text height=1.5 ex, text depth=0.25 ex] (19) at (0.75, 1) {Men};
		\node [style=none, text height=1.5 ex, text depth=0.25 ex] (20) at (3.75, 1) {built};
		\node [style=none, text height=1.5 ex, text depth=0.25 ex] (21) at (6.75, 1) {houses};
		\node [style=none, text height=1.5 ex, text depth=0.25 ex] (22) at (0.75, 0.25) {$W$};
		\node [style=none, text height=1.5 ex, text depth=0.25 ex] (23) at (3, 0.25) {$W$};
		\node [style=none, text height=1.5 ex, text depth=0.25 ex] (24) at (3.75, 0.25) {$W$};
		\node [style=none, text height=1.5 ex, text depth=0.25 ex] (25) at (4.5, 0.25) {$W$};
		\node [style=none, text height=1.5 ex, text depth=0.25 ex] (26) at (6.75, 0.25) {$W$};
		\node [style=none] (27) at (0.75, -0.25) {};
		\node [style=none] (28) at (3, -0.25) {};
		\node [style=none] (29) at (3.75, -0.25) {};
		\node [style=none] (30) at (4.5, -0.25) {};
		\node [style=none] (31) at (6.75, -0.25) {};
		\node [style=none] (32) at (3.75, -1.25) {};
	\end{pgfonlayer}
	\begin{pgfonlayer}{edgelayer}
		\draw [thick, looseness=0.00] (0.center) to (10.center);
		\draw [thick, bend right=90] (30.center) to (31.center);
		\draw [thick, looseness=0.00] (2.center) to (13.center);
		\draw [thick, looseness=0.00] (6.center) to (10.center);
		\draw [thick, looseness=0.00] (11.center) to (13.center);
		\draw [thick, looseness=0.00] (11.center) to (2.center);
		\draw [thick, looseness=0.00] (3.center) to (1.center);
		\draw [thick, looseness=0.00] (29.center) to (32.center);
		\draw [thick, looseness=0.00] (3.center) to (5.center);
		\draw [thick, looseness=0.00] (9.center) to (17.center);
		\draw [thick, looseness=0.00] (7.center) to (15.center);
		\draw [thick, looseness=0.00] (12.center) to (18.center);
		\draw [thick, bend right=90] (27.center) to (28.center);
		\draw [thick, looseness=0.00] (8.center) to (16.center);
		\draw [thick, looseness=0.00] (1.center) to (5.center);
		\draw [thick, looseness=0.00] (6.center) to (0.center);
		\draw [thick, looseness=0.00] (4.center) to (14.center);
	\end{pgfonlayer}
\end{tikzpicture}}

\end{center}

Computations with Frobenius algebras can  also be represented within the more general diagrammatic calculus of symmetric monoidal categories, referred to as \emph{string} diagrams, first formalized in \cite{Joyal}. Specifically, the  linear maps of the coalgebra and algebra are depicted by:

\vspace{0.2cm}
\begin{center}

\begin{tikzpicture}
	\begin{pgfonlayer}{nodelayer}
		\node [style=none] (0) at (-9.75, 3) {};
		\node [style=none] (1) at (-8.5, 3) {};
		\node [style=none] (2) at (-4, 3) {};
		\node [style=none] (3) at (-3, 3) {};
		\node [style=none] (4) at (-2.25, 3) {};
		\node [style=none] (5) at (-12, 2.25) {$(\sigma, \iota) = $};
		\node [fill=white, draw, thick, circle, minimum size=0.2 cm, style=none] (6) at (-9.75, 2.25) {};
		\node [style=none] (7) at (-5.75, 2.25) {$(\mu,\zeta)=$};
		\node [fill=white, draw, thick, circle, minimum size=0.2 cm, style=none] (8) at (-3.5, 2.25) {};
		\node [style=none] (9) at (-10.25, 1.5) {};
		\node [style=none] (10) at (-9.25, 1.5) {};
		\node [style=none] (11) at (-8.5, 1.5) {};
		\node [style=none] (12) at (-3.5, 1.5) {};
		\node [style=none] (13) at (-2.25, 1.5) {};
		\node [draw, circle, thick, minimum size=0.2 cm, fill=white, style=none] (14) at (-2.25, 3) {};
		\node [draw, circle, thick, minimum size=0.2 cm, fill=white, style=none] (15) at (-8.5, 1.5) {};
	\end{pgfonlayer}
	\begin{pgfonlayer}{edgelayer}
		\draw [thick, looseness=0.00] (11.center) to (1.center);
		\draw [thick, looseness=0.00] (8.center) to (12.center);
		\draw [thick, bend right=90, looseness=2.50] (2.center) to (3.center);
		\draw [thick, looseness=0.00] (4.center) to (13.center);
		\draw [thick, looseness=0.00] (6.center) to (0.center);
		\draw [thick, bend left=90, looseness=2.50] (9.center) to (10.center);
	\end{pgfonlayer}
\end{tikzpicture}}

\end{center} 
\vspace{0.2cm}

\noindent The \emph{Frobenius} condition is depicted by:

\vspace{-0.2cm}
\[

\begin{tikzpicture}
	\begin{pgfonlayer}{nodelayer}
		\node [style=none] (0) at (-1.25, 1.75) {$=$};
		\node [style=none] (1) at (0.75, 1.75) {$=$};
		\node [style=none] (2) at (0.25, 3) {};
		\node [style=none] (3) at (-0.75, 3) {};
		\node [draw, thick, style=none, minimum size=0.2 cm, circle, fill=white] (4) at (-0.25, 2.25) {};
		\node [style=none] (5) at (-0.75, 0) {};
		\node [style=none] (6) at (0.25, 0) {};
		\node [draw, thick, style=none, minimum size=0.2 cm, circle, fill=white] (7) at (-0.25, 0.75) {};
		\node [style=none] (8) at (-2.5, 3) {};
		\node [draw, thick, style=none, minimum size=0.2 cm, circle, fill=white] (9) at (-2.5, 2.25) {};
		\node [style=none] (10) at (-3, 1.5) {};
		\node [style=none] (11) at (-2, 1.5) {};
		\node [draw, thick, style=none, minimum size=0.2 cm, circle, fill=white] (12) at (-3.5, 0.75) {};
		\node [style=none] (13) at (-4, 1.5) {};
		\node [style=none] (14) at (-2, 0) {};
		\node [style=none] (15) at (-4, 3) {};
		\node [style=none] (16) at (-3.5, 0) {};
		\node [style=none] (17) at (1.5, 0) {};
		\node [style=none] (18) at (1.5, 0.75) {};
		\node [style=none] (19) at (2.5, 1.5) {};
		\node [style=none] (20) at (3.5, 1.5) {};
		\node [style=none] (21) at (1.5, 1.5) {};
		\node [draw, thick, style=none, minimum size=0.2 cm, circle, fill=white] (22) at (2, 2.25) {};
		\node [draw, thick, style=none, minimum size=0.2 cm, circle, fill=white] (23) at (3, 0.75) {};
		\node [style=none] (24) at (3, 0) {};
		\node [style=none] (25) at (2, 3) {};
		\node [style=none] (26) at (3.5, 3) {};
	\end{pgfonlayer}
	\begin{pgfonlayer}{edgelayer}
		\draw [thick, bend left=90, looseness=2.50] (2.center) to (3.center);
		\draw [style=thick, bend left=90, looseness=2.50] (5.center) to (6.center);
		\draw [style=thick] (4.center) to (7.center);
		\draw [style=thick] (8.center) to (9.center);
		\draw [style=thick, bend left=90, looseness=2.50] (10.center) to (11.center);
		\draw [style=thick, bend left=270, looseness=2.50] (13.center) to (10.center);
		\draw [style=thick] (11.center) to (14.center);
		\draw [style=thick] (15.center) to (13.center);
		\draw [style=thick] (12.center) to (16.center);
		\draw [style=thick] (17.center) to (18.center);
		\draw [style=thick, bend left=90, looseness=2.50] (21.center) to (19.center);
		\draw [style=thick, bend left=270, looseness=2.50] (19.center) to (20.center);
		\draw [style=thick] (23.center) to (24.center);
		\draw [style=thick] (25.center) to (22.center);
		\draw [style=thick] (21.center) to (18.center);
		\draw [style=thick] (26.center) to (20.center);
	\end{pgfonlayer}
\end{tikzpicture}}

\]

\noindent The commutativity conditions are shown as:

\begin{center}

\begin{tikzpicture}
	\begin{pgfonlayer}{nodelayer}
		\node [style=none] (0) at (-11.25, 3) {};
		\node [style=none] (1) at (-4, 1.75) {};
		\node [style=none] (2) at (-3, 1.75) {};
		\node [fill=white, draw, thick, circle, minimum size=0.2 cm, style=none] (3) at (-11.25, 2.25) {};
		\node [fill=white, draw, thick, circle, minimum size=0.2 cm, style=none] (4) at (-3.5, 1) {};
		\node [style=none] (5) at (-11.75, 1.5) {};
		\node [style=none] (6) at (-10.75, 1.5) {};
		\node [style=none] (7) at (-3.5, 0.25) {};
		\node [style=none] (8) at (-8.75, 3) {};
		\node [style=none] (9) at (-9.25, 1.5) {};
		\node [style=none] (10) at (-8.25, 1.5) {};
		\node [fill=white, draw, thick, circle, minimum size=0.2 cm, style=none] (11) at (-8.75, 2.25) {};
		\node [style=none] (12) at (-10, 1.5) {=};
		\node [style=none] (13) at (-1.5, 1.75) {};
		\node [fill=white, draw, thick, circle, minimum size=0.2 cm, style=none] (14) at (-1, 1) {};
		\node [style=none] (15) at (-1, 0.25) {};
		\node [style=none] (16) at (-0.5, 1.75) {};
		\node [style=none] (17) at (-2.25, 1.75) {=};
		\node [style=none] (18) at (-11.75, 0.25) {};
		\node [style=none] (19) at (-10.75, 0.25) {};
		\node [style=none] (20) at (-3, 1.75) {};
		\node [style=none] (21) at (-4, 1.75) {};
		\node [style=none] (22) at (-4, 3) {};
		\node [style=none] (23) at (-3, 3) {};
	\end{pgfonlayer}
	\begin{pgfonlayer}{edgelayer}
		\draw [thick, looseness=0.00] (4.center) to (7.center);
		\draw [thick, bend right=90, looseness=2.50] (1.center) to (2.center);
		\draw [thick, looseness=0.00] (3.center) to (0.center);
		\draw [thick, bend left=90, looseness=2.50] (5.center) to (6.center);
		\draw [thick, looseness=0.00] (11.center) to (8.center);
		\draw [thick, bend left=90, looseness=2.50] (9.center) to (10.center);
		\draw [thick, looseness=0.00] (14.center) to (15.center);
		\draw [thick, bend right=90, looseness=2.50] (13.center) to (16.center);
		\draw [thick, in=60, out=270, looseness=1.50] (6.center) to (18.center);
		\draw [thick, in=120, out=270, looseness=1.25] (5.center) to (19.center);
		\draw [thick, in=60, out=270, looseness=1.50] (23.center) to (21.center);
		\draw [thick, in=120, out=270, looseness=1.25] (22.center) to (20.center);
	\end{pgfonlayer}
\end{tikzpicture}}

\end{center} 

\noindent The isometry condition is depicted by:

\begin{center}

\begin{tikzpicture}
	\begin{pgfonlayer}{nodelayer}
		\node [style=none] (0) at (-9.75, 3) {};
		\node [style=none] (1) at (-7.25, 3) {};
		\node [style=none] (2) at (-10.25, 1.5) {};
		\node [style=none] (3) at (-9.25, 1.5) {};
		\node [fill=white, draw, thick, circle, minimum size=0.2 cm, style=none] (4) at (-9.75, 2.25) {};
		\node [fill=white, draw, thick, circle, minimum size=0.2 cm, style=none] (5) at (-9.75, 0.75) {};
		\node [style=none] (6) at (-10.25, 1.5) {};
		\node [style=none] (7) at (-9.25, 1.5) {};
		\node [style=none] (8) at (-7.25, 0) {};
		\node [style=none] (9) at (-9.75, 0) {};
		\node [style=none] (10) at (-8.25, 1.5) {=};
	\end{pgfonlayer}
	\begin{pgfonlayer}{edgelayer}
		\draw [thick, looseness=0.00] (8.center) to (1.center);
		\draw [thick, looseness=0.00] (5.center) to (9.center);
		\draw [thick, bend right=90, looseness=2.50] (2.center) to (3.center);
		\draw [thick, looseness=0.00] (4.center) to (0.center);
		\draw [thick, bend left=90, looseness=2.50] (6.center) to (7.center);
	\end{pgfonlayer}
\end{tikzpicture}}

\end{center} 

Finally, the Frobenius conditions guarantee that any diagram depicting  a Frobenius algebraic computation can be reduced to a normal form that only depends on the number of input and output wires of the nodes, provided that the diagram of computation is connected. This justifies depicting computations with Frobenius algebras as  \emph{spiders}, referring to the right hand side diagram below:

\vspace{-0.2cm}
\[

\begin{tikzpicture}
	\begin{pgfonlayer}{nodelayer}
		\node [style=none] (0) at (-4, 3.25) {};
		\node [style=none] (1) at (-3, 3.25) {};
		\node [draw, thick, style=none, minimum size=0.2 cm, circle, fill=white]  (2) at (-3.5, 2.5) {};
		\node [style=none] (3) at (-2.5, 2.5) {};
		\node [style=none] (4) at (-2.5, 3.25) {};
		\node[draw, thick, style=none, minimum size=0.2 cm, circle, fill=white]  (5) at (-3, 1.75) {};
		\node [style=none] (6) at (-2.5, 1.25) {$\ddots$};
		\node [style=none] (7) at (-2, 0.75) {};
		\node [style=none] (8) at (-1.25, 3.25) {};
		\node [style=none] (9) at (0, 0.5) {$=$};
		\node [style=none] (10) at (1.25, 1.75) {};
		\node [style=none] (11) at (3.75, 1.75) {};
		\node [draw, thick, style=none, minimum size=0.2 cm, circle, fill=white]  (12) at (2.5, 0.5) {};
		\node [style=none] (13) at (2, 1.75) {};
		\node [style=none] (14) at (3, 1.75) {$\cdots$};
		\node [style=none] (15) at (1.25, -0.75) {};
		\node [style=none] (16) at (3.75, -0.75) {};
		\node [style=none] (17) at (2, -0.75) {};
		\node [style=none] (18) at (3, -0.75) {$\cdots$};
		\node [style=none] (19) at (-4, -2.25) {};
		\node [style=none] (20) at (-3, -2.25) {};
		\node [draw, thick, style=none, minimum size=0.2 cm, circle, fill=white]  (21) at (-3.5, -1.5) {};
		\node [style=none] (22) at (-2.5, -1.5) {};
		\node [draw, thick, style=none, minimum size=0.2 cm, circle, fill=white] (23) at (-3, -0.75) {};
		\node [style=none] (24) at (-2.5, -0.25) {$\ddots$};
		\node [style=none] (25) at (-1.25, -2.25) {};
		\node [style=none] (26) at (-2, 0) {};
	\end{pgfonlayer}
	\begin{pgfonlayer}{edgelayer}
		\draw [style=thick, bend left=270, looseness=2.25] (0.center) to (1.center);
		\draw [style=thick, bend left=270, looseness=2.25] (2.center) to (3.center);
		\draw [style=thick] (4.center) to (3.center);
		\draw [style=thick, bend left=270, looseness=1.75] (10.center) to (11.center);
		\draw [style=thick] (13.center) to (12.center);
		\draw [style=thick, bend left=90, looseness=1.50] (15.center) to (16.center);
		\draw [style=thick] (17.center) to (12.center);
		\draw [style=thick, bend left=90, looseness=2.50] (19.center) to (20.center);
		\draw [style=thick, bend left=90, looseness=2.50] (21.center) to (22.center);
		\draw [style=thick, in=90, out=-60] (26.center) to (25.center);
		\draw [style=thick, in=45, out=-90] (8.center) to (7.center);
	\end{pgfonlayer}
\end{tikzpicture}}

\]

For an informal introduction to compact closed categories, Frobenius algebras, and their diagrammatic calculi, see \cite{CoeckePaq}.

%
%

\section{Building Tensors for Words with Complex Types}
\label{sec:frobop}

The type-logical models of meaning  treat words with complex types as predicates. In a matrix calculus, predicates can be modeled as matrices (or equivalently, linear maps),  over the semiring of booleans. In vector spaces over reals, one can extend these 0/1 entries to real numbers and model words with complex types as weighted predicates. These are predicates that not only tell us which instantiations of their arguments  are related to each other, but also that to what extent these are related to each other. For instance, a transitive verb is a binary predicate that, in the type-logical model, tells us which noun phrases are related to other noun phrases. In a vector space model, the verb becomes a linear map that moreover tells us to what extent these noun phrases are related to each other. 

Building such linear maps from a corpus turns out to be a non-trivial task. In previous work \cite{Grefenetal,GrefenSadr1} we argue that such a linear map can be constructed by taking the sum of the tensor products of the vectors/linear maps of its arguments. For instance, the linear map representing an $n$-ary predicate $p$  with arguments $a_1$ to $a_n$ is $\sum_{i} \overline{a}_1 \otimes \cdots \otimes \overline{a}_n$, where $\overline{a}_j$ is the vector/linear map associated to the argument $a_j$ and the index $i$ counts the number of times each word $a_j$ has appeared in the corpus as the argument of $p$. Following this method, the linear maps corresponding to the predicates of our simple grammar are as follows:

\begin{center}
\begin{tabular}{c|c|c}
intransitive verb \ & \ transitive verb \ & \ adjective\\
\hline
&&\\
$\sum_i \ov{sbj}_i$ \ & \ $\sum\limits_{i}(\ov{sbj}_i\otimes \ov{obj}_i)$ \ & \ $\sum_i \ov{noun}_i$\\
&&\\
\end{tabular}
\end{center}


\noindent There is a  problem: this method provides us with a linear map in a space whose tensor rank is one less than the rank of the space needed by Definition(*). For instance, the linear map of the transitive verb ends up being in  $W \otimes W$, but we need a linear map in $W \otimes W \otimes W$. This problem is overcome by using the operations of a Frobenius algebra over vector spaces. We use the pictorial calculi of the compact closed categories and Frobenius algebras to depict the resulting linear maps and sentence vectors. 

\subsection{Adjectives and Intransitive Verbs} 

The linear maps of adjectives and intransitive verbs are  elements of $W$. In order to encode them in $W \otimes W$, we use the Frobenius $\sigma$ operation and obtain the following linear map:
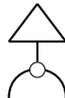
\begin{center}

\begin{tikzpicture}
	\begin{pgfonlayer}{nodelayer}
		\node [draw, circle, minimum size=0.2 cm, fill=white, style=none] (0) at (2.75, 1.25) {};
		\node [style=none] (1) at (2, 0.5) {};
		\node [style=none] (2) at (3.5, 0.5) {};
		\node [style=none] (3) at (2.75, 3) {};
		\node [style=none] (4) at (2, 2) {};
		\node [style=none] (5) at (3.5, 2) {};
		\node [style=none] (6) at (2.75, 2) {};
	\end{pgfonlayer}
	\begin{pgfonlayer}{edgelayer}
		\draw [style=thick, bend left=90, looseness=1.75] (1.center) to (2.center);
		\draw [style=thick] (4.center) to (5.center);
		\draw [style=thick] (3.center) to (4.center);
		\draw [style=thick] (3.center) to (5.center);
		\draw [style=thick] (6.center) to (0);
	\end{pgfonlayer}
\end{tikzpicture}}

\end{center}
For the intransitive verb, when substituted in Definition(*), that is, when applied to its subject, the above will result in the left hand side vector below, which is then normalized to the right hand side vector. 
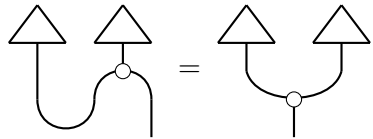
\begin{center}

\begin{tikzpicture}
	\begin{pgfonlayer}{nodelayer}
		\node [style=none] (0) at (4, 1.25) {};
		\node [draw, circle, minimum size=0.2 cm, fill=white, style=none] (1) at (2.75, 0.5) {};
		\node [style=none] (2) at (0.75, 2) {};
		\node [style=none] (3) at (1.5, 2) {};
		\node [style=none] (4) at (2.25, 2) {};
		\node [style=none] (5) at (1.5, 3) {};
		\node [style=none] (6) at (1.5, 1.25) {};
		\node [style=none] (7) at (2.75, -0.5) {};
		\node [style=none] (8) at (4, 3) {};
		\node [style=none] (9) at (3.25, 2) {};
		\node [style=none] (10) at (4.75, 2) {};
		\node [style=none] (11) at (4, 2) {};
		\node [style=none] (12) at (-1, 0.5) {};
		\node [style=none] (13) at (-4, 3) {};
		\node [style=none] (14) at (-4, 0.5) {};
		\node [style=none] (15) at (-4, 2) {};
		\node [draw, circle, minimum size=0.2 cm, fill=white, style=none] (16) at (-1.75, 1.25) {};
		\node [style=none] (17) at (-2.5, 0.5) {};
		\node [style=none] (18) at (-1.75, 2) {};
		\node [style=none] (19) at (-4.75, 2) {};
		\node [style=none] (20) at (-3.25, 2) {};
		\node [style=none] (21) at (-1, -0.5) {};
		\node [style=none] (22) at (-2.5, 2) {};
		\node [style=none] (23) at (-1.75, 3) {};
		\node [style=none] (24) at (-1, 2) {};
		\node [style=none] (25) at (0, 1.25) {$=$};
	\end{pgfonlayer}
	\begin{pgfonlayer}{edgelayer}
		\draw [thick, looseness=0.00] (2.center) to (5.center);
		\draw [thick, looseness=0.00] (5.center) to (4.center);
		\draw [thick, looseness=0.00] (2.center) to (4.center);
		\draw [style=thick] (3.center) to (6.center);
		\draw [style=thick] (1.center) to (7.center);
		\draw [style=thick] (9.center) to (10.center);
		\draw [style=thick] (8.center) to (9.center);
		\draw [style=thick] (8.center) to (10.center);
		\draw [style=thick] (11.center) to (0.center);
		\draw [style=thick, bend right=75] (6.center) to (0.center);
		\draw [style=thick, bend left=90, looseness=1.75] (17.center) to (12.center);
		\draw [thick, looseness=0.00] (19.center) to (13.center);
		\draw [thick, looseness=0.00] (13.center) to (20.center);
		\draw [thick, looseness=0.00] (19.center) to (20.center);
		\draw [style=thick] (15.center) to (14.center);
		\draw [style=thick, bend right=90, looseness=1.75] (14.center) to (17.center);
		\draw [style=thick] (12.center) to (21.center);
		\draw [style=thick] (22.center) to (24.center);
		\draw [style=thick] (23.center) to (22.center);
		\draw [style=thick] (23.center) to (24.center);
		\draw [style=thick] (18.center) to (16.center);
	\end{pgfonlayer}
\end{tikzpicture}}

\end{center}
When an adjective is applied to a noun, the order of the  above application is swapped: the triangle  of the adjective will change place with the triangle of the subject of the intransitive verb. 

\subsection{Transitive Verbs}

The linear map of a transitive verb is an element of $W \otimes W$; this has to be encoded in $W \otimes W \otimes W$.  We face a few options here, which geometrically speaking provide us with different ways of ``diagonally'' placing a plane into a cube.   

\paragraph{{\sc \textbf{CpSbj}}} The first option is to copy the ``row'' dimension of the linear map corresponding to the verb; this dimension encodes the information of the  subjects of the verb from the corpus. In the left hand side diagram below we see how $\sigma$ transforms the verb in this way. Once substituted in Definition(*), we obtain the diagram in the right hand side:

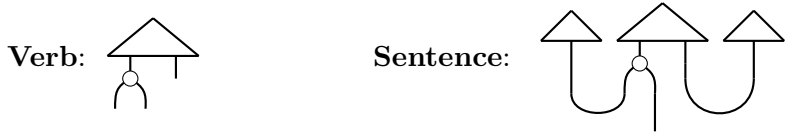
\begin{center}

\begin{tikzpicture}[scale=0.8]
	\begin{pgfonlayer}{nodelayer}
		\node [style=none] (0) at (8.75, 3.25) {};
		\node [style=none] (1) at (5.75, 3) {};
		\node [style=none] (2) at (11.75, 3) {};
		\node [style=none] (3) at (-8, 2.75) {};
		\node [style=none] (4) at (4.75, 2) {};
		\node [style=none] (5) at (5.75, 2) {};
		\node [style=none] (6) at (6.75, 2) {};
		\node [style=none] (7) at (7.25, 2) {};
		\node [style=none] (8) at (8, 2) {};
		\node [style=none] (9) at (9.5, 2) {};
		\node [style=none] (10) at (10.25, 2) {};
		\node [style=none] (11) at (10.75, 2) {};
		\node [style=none] (12) at (11.75, 2) {};
		\node [style=none] (13) at (12.75, 2) {};
		\node [style=none] (14) at (-9.5, 1.5) {};
		\node [style=none] (15) at (-8.75, 1.5) {};
		\node [style=none] (16) at (-7.25, 1.5) {};
		\node [style=none] (17) at (-6.5, 1.5) {};
		\node [draw, circle, minimum size=0.2 cm, fill=white, style=none] (18) at (8, 1.25) {};
		\node [style=none] (19) at (8, 1.25) {};
		\node [style=none] (20) at (9.5, 1.25) {};
		\node [style=none] (21) at (9.5, 0.75) {};
		\node [style=none] (22) at (11.75, 0.75) {};
		\node [draw, circle, minimum size=0.2 cm, fill=white, style=none] (23) at (-8.75, 0.75) {};
		\node [style=none] (24) at (-8.75, 0.75) {};
		\node [style=none] (25) at (-7.25, 0.75) {};
		\node [style=none] (26) at (5.75, 0.25) {};
		\node [style=none] (27) at (7.5, 0.25) {};
		\node [style=none] (28) at (8.5, 0.25) {};
		\node [style=none] (29) at (-9.25, -0.25) {};
		\node [style=none] (30) at (-8.25, -0.25) {};
		\node [style=none] (31) at (8.5, -1) {};
		\node [style=none] (32) at (-11.5, 1.5) {{\bf Verb}:};
		\node [style=none] (33) at (1.5, 1.5) {{\bf Sentence}:};
	\end{pgfonlayer}
	\begin{pgfonlayer}{edgelayer}
		\draw [thick, looseness=0.00] (1.center) to (6.center);
		\draw [thick, looseness=0.00] (3.center) to (17.center);
		\draw [thick, bend left=270, looseness=1.25] (26.center) to (27.center);
		\draw [thick, looseness=0.00] (15.center) to (24.center);
		\draw [thick, looseness=0.00] (2.center) to (13.center);
		\draw [thick, looseness=0.00] (7.center) to (0.center);
		\draw [thick, looseness=0.00] (7.center) to (10.center);
		\draw [thick, looseness=0.00] (8.center) to (19.center);
		\draw [thick, bend left, looseness=1.25] (19.center) to (28.center);
		\draw [thick, looseness=0.00] (9.center) to (20.center);
		\draw [thick, looseness=0.00] (4.center) to (1.center);
		\draw [thick, in=90, out=210, looseness=1.25] (24.center) to (29.center);
		\draw [thick, looseness=0.00] (5.center) to (26.center);
		\draw [thick, looseness=0.00] (16.center) to (25.center);
		\draw [thick, looseness=0.00] (0.center) to (10.center);
		\draw [thick, in=90, out=-90] (20.center) to (21.center);
		\draw [thick, bend left, looseness=1.25] (24.center) to (30.center);
		\draw [thick, in=90, out=210, looseness=1.25] (19.center) to (27.center);
		\draw [thick, looseness=0.00] (12.center) to (22.center);
		\draw [thick, looseness=0.00] (14.center) to (17.center);
		\draw [thick, looseness=0.00] (4.center) to (6.center);
		\draw [thick, looseness=0.00] (11.center) to (2.center);
		\draw [thick, bend right=90, looseness=1.75] (21.center) to (22.center);
		\draw [thick, looseness=0.00] (14.center) to (3.center);
		\draw [thick, looseness=0.00] (28.center) to (31.center);
		\draw [thick, looseness=0.00] (11.center) to (13.center);
	\end{pgfonlayer}
\end{tikzpicture}}

\end{center}
\vspace{+0.2cm}

\noindent In this case, the $\sigma$ map transforms the matrix of the verb as follows:

\vspace{-0.2cm}
\[
 \sigma:: \sum\limits_{ij} c_{ij} (\ov{n_i} \otimes \ov{n_j}) \quad \mapsto  \quad 
 \sum\limits_{ij} c_{ij}(\ov{n_i} \otimes \ov{n_i} \otimes \ov{n_j})
\]

\paragraph{{\sc \textbf{CpObj}}} Our other option is to copy the ``column'' dimension of the matrix, which encodes the information about the  objects of the verb from the corpus: 

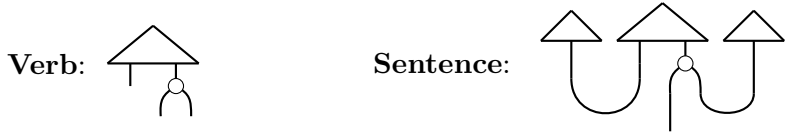
\begin{center}

\begin{tikzpicture}[scale=0.8]
	\begin{pgfonlayer}{nodelayer}
		\node [style=none] (0) at (8.75, 3.25) {};
		\node [style=none] (1) at (5.75, 3) {};
		\node [style=none] (2) at (11.75, 3) {};
		\node [style=none] (3) at (-8, 2.5) {};
		\node [style=none] (4) at (4.75, 2) {};
		\node [style=none] (5) at (5.75, 2) {};
		\node [style=none] (6) at (6.75, 2) {};
		\node [style=none] (7) at (7.25, 2) {};
		\node [style=none] (8) at (8, 2) {};
		\node [style=none] (9) at (9.5, 2) {};
		\node [style=none] (10) at (10.25, 2) {};
		\node [style=none] (11) at (10.75, 2) {};
		\node [style=none] (12) at (11.75, 2) {};
		\node [style=none] (13) at (12.75, 2) {};
		\node [style=none] (14) at (-9.5, 1.25) {};
		\node [style=none] (15) at (-8.75, 1.25) {};
		\node [style=none] (16) at (-7.25, 1.25) {};
		\node [style=none] (17) at (-6.5, 1.25) {};
		\node [style=none] (18) at (8, 1.25) {};
		\node [draw, circle, minimum size=0.2 cm, fill=white, style=none] (19) at (9.5, 1.25) {};
		\node [style=none] (20) at (9.5, 1.25) {};
		\node [style=none] (21) at (5.75, 0.75) {};
		\node [style=none] (22) at (8, 0.75) {};
		\node [style=none] (23) at (-8.75, 0.5) {};
		\node [draw, circle, minimum size=0.2 cm, fill=white, style=none] (24) at (-7.25, 0.5) {};
		\node [style=none] (25) at (-7.25, 0.5) {};
		\node [style=none] (26) at (9, 0.25) {};
		\node [style=none] (27) at (10, 0.25) {};
		\node [style=none] (28) at (11.75, 0.25) {};
		\node [style=none] (29) at (-7.75, -0.5) {};
		\node [style=none] (30) at (-6.75, -0.5) {};
		\node [style=none] (31) at (9, -1) {};
		\node [style=none] (32) at (-11.5, 1.25) {{\bf Verb}:};
		\node [style=none] (33) at (1.5, 1.25) {{\bf Sentence}:};
	\end{pgfonlayer}
	\begin{pgfonlayer}{edgelayer}
		\draw [thick, in=90, out=-30, looseness=1.25] (20.center) to (27.center);
		\draw [thick, looseness=0.00] (13.center) to (2.center);
		\draw [thick, looseness=0.00] (0.center) to (7.center);
		\draw [thick, looseness=0.00] (16.center) to (25.center);
		\draw [thick, looseness=0.00] (6.center) to (1.center);
		\draw [thick, bend right, looseness=1.25] (25.center) to (29.center);
		\draw [thick, looseness=0.00] (13.center) to (11.center);
		\draw [thick, looseness=0.00] (17.center) to (14.center);
		\draw [thick, looseness=0.00] (3.center) to (14.center);
		\draw [thick, looseness=0.00] (2.center) to (11.center);
		\draw [thick, looseness=0.00] (1.center) to (4.center);
		\draw [thick, looseness=0.00] (17.center) to (3.center);
		\draw [thick, bend right=270, looseness=1.25] (28.center) to (27.center);
		\draw [thick, looseness=0.00] (9.center) to (20.center);
		\draw [thick, looseness=0.00] (10.center) to (0.center);
		\draw [thick, looseness=0.00] (10.center) to (7.center);
		\draw [thick, looseness=0.00] (15.center) to (23.center);
		\draw [thick, bend right, looseness=1.25] (20.center) to (26.center);
		\draw [thick, looseness=0.00] (5.center) to (21.center);
		\draw [thick, looseness=0.00] (6.center) to (4.center);
		\draw [thick, in=90, out=-90] (18.center) to (22.center);
		\draw [thick, bend left=90, looseness=1.75] (22.center) to (21.center);
		\draw [thick, looseness=0.00] (26.center) to (31.center);
		\draw [thick, looseness=0.00] (12.center) to (28.center);
		\draw [thick, looseness=0.00] (8.center) to (18.center);
		\draw [thick, in=90, out=-30, looseness=1.25] (25.center) to (30.center);
	\end{pgfonlayer}
\end{tikzpicture}}

\end{center}
\vspace{+0.2cm}

\noindent Now the $\sigma$-map does the following transformation:

\vspace{-0.2cm}
\[
 \sigma:: \sum\limits_{ij} c_{ij} (\ov{n_i} \otimes \ov{n_j}) \quad \mapsto\quad
 \sum\limits_{ij} c_{ij}(\ov{n_i} \otimes \ov{n_j} \otimes \ov{n_j})
\]

The diagrams above simplify the calculations involved, since they suggest a closed form formula for each case. Taking as an example the diagram of the copy-subject method, we see that: (a) the object interacts with the verb; (b) the result of this interaction serves as input for the $\sigma$ map; (c) one wire of the output of $\sigma$ interacts with the subject, while the other branch delivers the result. Linear algebraically, this corresponds to the computation $\sigma(\overline{verb} \times \ov{obj})^{\mathsf{T}} \times \ov{sbj}$, which expresses the fact that the meaning of a sentence is obtained by first applying the meaning of the verb to the meaning of the object, then applying the ($\sigma$ version of the) result  to the meaning of the subject. This computation results in the Equation \ref{equ:clfrm_sbj} below:

\vspace{-0.2cm}
\begin{equation}
\label{equ:clfrm_sbj}
  \ov{sbj~verb~obj} = \ov{sbj} \odot (\overline{verb} \times \ov{obj})
\end{equation}

\noindent This order of application is the exact same way formalized in the generative rules of the language. On the contrary, the meaning of a transitive sentence for the copy-object results is given by Equation \ref{equ:clfrm_obj} below, which expresses the fact that the meaning of a sentence is  obtained by first applying the (transposed) meaning of the verb to the meaning of the subject and then applying the result to the meaning of the object:

\vspace{-0.2cm}
\begin{equation}
\label{equ:clfrm_obj}
  \ov{sbj~verb~obj} = \ov{obj} \odot (\overline{verb}^{\mathsf{T}} \times \ov{sbj})
\end{equation}


Note that equipped with the above closed forms we do not need to create or manipulate rank-3 tensors at any point of the computation, something that would cause high computational overhead.  

Purely syntactically speaking, in a pregroup grammar the order of application of a transitive verb does not matter: it is applied to its subject and object in parallel. Semantically, as originated in the work of Montague \cite{Montague}, a transitive verb is first applied to its object  and  then  to its subject. In the more modern approaches to semantics via logical grammars, this order is some times based on the choice of the specific verb \cite{Lamarche}. Our work in this paper is more inline with the latter approach, where for the specific task of disambiguating the verbs of our dataset, first applying the verb to the subject then to the object seems to provide better experimental results.  According to our general theoretical setting,  the linear map corresponding to the transitive verb should be a rank-3 tensor, but at the moment,  apart from work in progress which tries to conjoin efforts with Machine Learning to directly build these as rank-3 tensors, we do not have the technology to do other than described in this paper. However, in the ideal case that the linear maps of words are already in the spaces allocated to them by the theory, these choices will not arise, as the compact nature of the matrix calculus implies that the application can be done in  parallel in all the cases that parallel applications are prescribed by the syntax. From a linear-algebra perspective, fully populated rank-3 tensors for verbs satisfy the following equality:

\[
  \overline{subj~verb~obj} = (\overline{verb} \times \ov{obj})^{\mathsf{T}} \times \ov{subj} = (\overline{verb}^{\mathsf{T}} \times \ov{subj}) \times \ov{obj}
\]

\noindent which shows that the order of application does not actually play a role. 

%
%
%

\paragraph{\sc  \textbf{MixCpDl}} We can also use a mixture of $\sigma$ and $\mu$ maps. There are three reasonable options here, all of which start by applying two $\sigma$'s to the two  wires of the linear map of the verb (that is, one for each of the dimensions). Then one can either apply a $\iota$ to one of the copies of the first wire, or a $\iota$ to one of the copies of the second wire. These two options are depicted as follows:

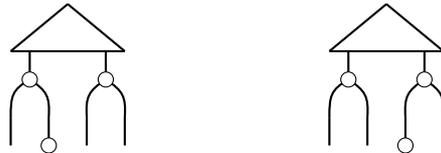
\vspace{-0.2cm}
\begin{center}

\begin{tikzpicture}
	\begin{pgfonlayer}{nodelayer}
		\node [style=none] (0) at (3.75, 3.25) {};
		\node [style=none] (1) at (2.25, 2) {};
		\node [style=none] (2) at (2.75, 2) {};
		\node [style=none] (3) at (4.75, 2) {};
		\node [style=none] (4) at (5.25, 2) {};
		\node [draw, circle, minimum size=0.2 cm, fill=white, style=none] (5) at (2.75, 1.25) {};
		\node [style=none] (6) at (2.75, 1.25) {};
		\node [style=none] (7) at (4.75, 1.25) {};
		\node [draw, circle, minimum size=0.2 cm, fill=white, style=none] (8) at (4.75, 1.25) {};
		\node [style=none] (9) at (2.25, 0.25) {};
		\node [style=none] (10) at (3.25, 0.25) {};
		\node [style=none] (11) at (4.25, 0.25) {};
		\node [style=none] (12) at (5.25, 0.25) {};
		\node [style=none] (13) at (3.25, 0.25) {};
		\node [draw, circle, minimum size=0.2 cm, fill=white, style=none] (14) at (3.25, -0.5) {};
		\node [style=none] (15) at (2.25, -0.5) {};
		\node [style=none] (16) at (4.25, -0.5) {};
		\node [style=none] (17) at (5.25, -0.5) {};
	\end{pgfonlayer}
	\begin{pgfonlayer}{edgelayer}
		\draw [thick, looseness=0.00] (1.center) to (0.center);
		\draw [thick, in=90, out=210, looseness=1.25] (6.center) to (9.center);
		\draw [thick, looseness=0.00] (2.center) to (6.center);
		\draw [thick, bend right, looseness=1.25] (7.center) to (11.center);
		\draw [thick, looseness=0.00] (1.center) to (4.center);
		\draw [thick, bend left, looseness=1.25] (6.center) to (10.center);
		\draw [thick, looseness=0.00] (3.center) to (7.center);
		\draw [thick, bend left, looseness=1.25] (7.center) to (12.center);
		\draw [thick, looseness=0.00] (0.center) to (4.center);
		\draw [style=thick] (13.center) to (14.center);
		\draw [style=thick] (9.center) to (15.center);
		\draw [style=thick] (11.center) to (16.center);
		\draw [style=thick] (12.center) to (17.center);
	\end{pgfonlayer}
\end{tikzpicture}}

\hspace{2cm}

\begin{tikzpicture}
	\begin{pgfonlayer}{nodelayer}
		\node [style=none] (0) at (3.75, 3.25) {};
		\node [style=none] (1) at (2.25, 2) {};
		\node [style=none] (2) at (2.75, 2) {};
		\node [style=none] (3) at (4.75, 2) {};
		\node [style=none] (4) at (5.25, 2) {};
		\node [draw, circle, minimum size=0.2 cm, fill=white, style=none] (5) at (2.75, 1.25) {};
		\node [style=none] (6) at (2.75, 1.25) {};
		\node [style=none] (7) at (4.75, 1.25) {};
		\node [draw, circle, minimum size=0.2 cm, fill=white, style=none] (8) at (4.75, 1.25) {};
		\node [style=none] (9) at (2.25, 0.25) {};
		\node [style=none] (10) at (3.25, 0.25) {};
		\node [style=none] (11) at (4.25, 0.25) {};
		\node [style=none] (12) at (5.25, 0.25) {};
		\node [style=none] (13) at (3.25, 0.25) {};
		\node [draw, circle, minimum size=0.2 cm, fill=white, style=none] (14) at (4.25, -0.5) {};
		\node [style=none] (15) at (2.25, -0.5) {};
		\node [style=none] (16) at (3.25, -0.5) {};
		\node [style=none] (17) at (5.25, -0.5) {};
	\end{pgfonlayer}
	\begin{pgfonlayer}{edgelayer}
		\draw [thick, looseness=0.00] (1.center) to (0.center);
		\draw [thick, in=90, out=210, looseness=1.25] (6.center) to (9.center);
		\draw [thick, looseness=0.00] (2.center) to (6.center);
		\draw [thick, bend right, looseness=1.25] (7.center) to (11.center);
		\draw [thick, looseness=0.00] (1.center) to (4.center);
		\draw [thick, bend left, looseness=1.25] (6.center) to (10.center);
		\draw [thick, looseness=0.00] (3.center) to (7.center);
		\draw [thick, bend left, looseness=1.25] (7.center) to (12.center);
		\draw [thick, looseness=0.00] (0.center) to (4.center);
		\draw [style=thick] (11.center) to (14.center);
		\draw [style=thick] (9.center) to (15.center);
		\draw [style=thick] (13.center) to (16.center);
		\draw [style=thick] (12.center) to (17.center);
	\end{pgfonlayer}
\end{tikzpicture}}

\end{center}

\vspace{-0.2cm}
\noindent 
The first diagram  has the same normal form as the copy-subject option, and the second one has the same normal form as the copy-object option. 


Finally, one can apply a $\mu$ to one wire from each of the copied wires of the verb, the result of which is  depicted in the following left hand side diagram. When substituted in Definition(*), we obtain the following right hand side diagram for the meaning of the transitive sentence:

\vspace{-0.2cm}
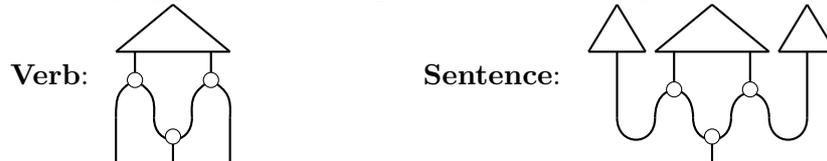
\begin{center}
 {\bf Verb}: 
\begin{tikzpicture}
	\begin{pgfonlayer}{nodelayer}
		\node [style=none] (0) at (3.75, 3.25) {};
		\node [style=none] (1) at (2.25, 2) {};
		\node [style=none] (2) at (2.75, 2) {};
		\node [style=none] (3) at (4.75, 2) {};
		\node [style=none] (4) at (5.25, 2) {};
		\node [draw, circle, minimum size=0.2 cm, fill=white, style=none] (5) at (2.75, 1.25) {};
		\node [style=none] (6) at (2.75, 1.25) {};
		\node [style=none] (7) at (4.75, 1.25) {};
		\node [draw, circle, minimum size=0.2 cm, fill=white, style=none] (8) at (4.75, 1.25) {};
		\node [style=none] (9) at (2.25, 0.25) {};
		\node [style=none] (10) at (3.25, 0.25) {};
		\node [style=none] (11) at (4.25, 0.25) {};
		\node [style=none] (12) at (5.25, 0.25) {};
		\node [draw, circle, minimum size=0.2 cm, fill=white, style=none] (13) at (3.75, -0.25) {};
		\node [style=none] (14) at (2.25, -1) {};
		\node [style=none] (15) at (3.75, -1) {};
		\node [style=none] (16) at (5.25, -1) {};
	\end{pgfonlayer}
	\begin{pgfonlayer}{edgelayer}
		\draw [thick, looseness=0.00] (12.center) to (16.center);
		\draw [thick, looseness=0.00] (1.center) to (0.center);
		\draw [thick, looseness=0.00] (9.center) to (14.center);
		\draw [thick, in=90, out=210, looseness=1.25] (6.center) to (9.center);
		\draw [thick, bend left=270, looseness=2.00] (10.center) to (11.center);
		\draw [thick, looseness=0.00] (13.center) to (15.center);
		\draw [thick, looseness=0.00] (2.center) to (6.center);
		\draw [thick, bend right, looseness=1.25] (7.center) to (11.center);
		\draw [thick, looseness=0.00] (1.center) to (4.center);
		\draw [thick, bend left, looseness=1.25] (6.center) to (10.center);
		\draw [thick, looseness=0.00] (3.center) to (7.center);
		\draw [thick, bend left, looseness=1.25] (7.center) to (12.center);
		\draw [thick, looseness=0.00] (0.center) to (4.center);
	\end{pgfonlayer}
\end{tikzpicture}}

 \hspace{2cm}
 {\bf Sentence}:  
\begin{tikzpicture}
	\begin{pgfonlayer}{nodelayer}
		\node [style=none] (0) at (-2, 3.25) {};
		\node [style=none] (1) at (-3.5, 2) {};
		\node [style=none] (2) at (-3, 2) {};
		\node [style=none] (3) at (-1, 2) {};
		\node [style=none] (4) at (-0.5, 2) {};
		\node  [draw, circle, minimum size=0.2 cm, fill=white, style=none] (5) at (-3, 1) {};
		\node [draw, circle, minimum size=0.2 cm, fill=white, style=none] (6) at (-1, 1) {};
		\node [style=none] (7) at (-3.5, 0.25) {};
		\node [style=none] (8) at (-2.5, 0.25) {};
		\node [style=none] (9) at (-1.5, 0.25) {};
		\node [style=none] (10) at (-0.5, 0.25) {};
		\node [draw, circle, minimum size=0.2 cm, fill=white, style=none] (11) at (-2, -0.25) {};
		\node [style=none] (12) at (-2, -1) {};
		\node [style=none] (13) at (-5.25, 2) {};
		\node [style=none] (14) at (-4.5, 3.25) {};
		\node [style=none] (15) at (-3.75, 2) {};
		\node [style=none] (16) at (-0.25, 2) {};
		\node [style=none] (17) at (1.25, 2) {};
		\node [style=none] (18) at (0.5, 3.25) {};
		\node [style=none] (19) at (-4.5, 0.25) {};
		\node [style=none] (20) at (0.5, 0.25) {};
		\node [style=none] (21) at (-4.5, 2) {};
		\node [style=none] (22) at (0.5, 2) {};
	\end{pgfonlayer}
	\begin{pgfonlayer}{edgelayer}
		\draw [thick, looseness=0.00] (1.center) to (0.center);
		\draw [thick, bend left=270, looseness=2.00] (8.center) to (9.center);
		\draw [thick, looseness=0.00] (11.center) to (12.center);
		\draw [thick, looseness=0.00] (2.center) to (5.center);
		\draw [thick, looseness=0.00] (1.center) to (4.center);
		\draw [thick, looseness=0.00] (3.center) to (6.center);
		\draw [thick, looseness=0.00] (0.center) to (4.center);
		\draw [style=thick] (14.center) to (13.center);
		\draw [style=thick] (13.center) to (15.center);
		\draw [style=thick] (14.center) to (15.center);
		\draw [style=thick] (18.center) to (16.center);
		\draw [style=thick] (16.center) to (17.center);
		\draw [style=thick] (18.center) to (17.center);
		\draw [style=thick] (21.center) to (19.center);
		\draw [style=thick] (22.center) to (20.center);
		\draw [style=thick, bend left=270, looseness=2.00] (19.center) to (7.center);
		\draw [style=thick, bend left=270, looseness=2.00] (10.center) to (20.center);
		\draw [style=thick, bend left=90, looseness=2.50] (7.center) to (8.center);
		\draw [style=thick, bend left=90, looseness=2.25] (9.center) to (10.center);
	\end{pgfonlayer}
\end{tikzpicture}}

\end{center}

%
%
\vspace{-0.2cm}
\noindent The normal form of the diagram of the sentence  is obtained by  collapsing the three dots and yanking the corresponding wires, resulting in  the following diagram:

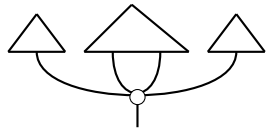
\vspace{-0.2cm}
\begin{center}
 
\begin{tikzpicture}
	\begin{pgfonlayer}{nodelayer}
		\node [style=none] (0) at (2, 3) {};
		\node [style=none] (1) at (4.5, 3.25) {};
		\node [style=none] (2) at (7.25, 3) {};
		\node [style=none] (3) at (1.25, 2) {};
		\node [style=none] (4) at (2, 2) {};
		\node [style=none] (5) at (2.75, 2) {};
		\node [style=none] (6) at (3.25, 2) {};
		\node [style=none] (7) at (4, 2) {};
		\node [style=none] (8) at (5.25, 2) {};
		\node [style=none] (9) at (6, 2) {};
		\node [style=none] (10) at (6.5, 2) {};
		\node [style=none] (11) at (7.25, 2) {};
		\node [style=none] (12) at (8, 2) {};
		\node [draw, circle, minimum size=0.2 cm, fill=white, style=none] (13) at (4.65, 0.8) {};
		\node [style=none] (14) at (4.65, -0) {};
	\end{pgfonlayer}
	\begin{pgfonlayer}{edgelayer}
		\draw [thick, bend right=90, looseness=0.75] (4.center) to (11.center);
		\draw [thick, bend right=90, looseness=3.00] (7.center) to (8.center);
		\draw [thick, looseness=0.00] (2.center) to (12.center);
		\draw [thick, looseness=0.00] (3.center) to (0.center);
		\draw [thick, looseness=0.00] (10.center) to (2.center);
		\draw [thick, looseness=0.00] (0.center) to (5.center);
		\draw [thick, looseness=0.00] (13.center) to (14.center);
		\draw [thick, looseness=0.00] (10.center) to (12.center);
		\draw [thick, looseness=0.00] (3.center) to (5.center);
		\draw [thick, looseness=0.00] (9.center) to (6.center);
		\draw [thick, looseness=0.00] (1.center) to (6.center);
		\draw [thick, looseness=0.00] (1.center) to (9.center);
	\end{pgfonlayer}
\end{tikzpicture}}

\end{center}

\noindent
Linear algebraically, the spider form of the verb is equivalent to $\sum_i (\ov{sbj_i} \odot \ov{obj_i})$. A verb obtained in this way will only relate the properties of its subjects and objects on identical bases and there will be no interaction of properties across bases. For instance, for a certain verb $v$, this construction will result in a vector that only encodes to what extent $v$ has related subjects and objects with property $\ov{w_1}$, and has no information about to what extent $v$ has related subjects with property $\ov{w_1}$ to objects with property $\ov{w_2}$. The closed form of the above diagram is:
\[
\ov{sbj} \odot \big(\sum_i (\ov{sbj_i} \odot \ov{obj_i})\big) \odot \ov{obj}
\]

\subsection{Encoding the Existing Non-Predicative Models}

Apart from the predicative way of encoding meanings of words with complex types, there exists two other approaches in the literature, who simply work with the context vectors of such words \cite{Lapata,GrefenSadr2}. These two approaches are representable in our setting using the Frobenius operations. 

\paragraph{{\sc \textbf{Multp}}} To represent the model of \cite{Lapata} in our setting, in which the meaning of a sentence is simply the point-wise multiplication of the context vectors of the words, we start from the context vector of the verb, denoted by $\ov{verb}$, and apply three  $\sigma$'s  and  then one $\mu$ to it. The result is depicted in the left hand side diagram below; once this verb is substituted in Definition(*), we obtain the right hand side diagram below as the meaning of a transitive sentence:

\vspace{-0.2cm}
\begin{center}
\textbf{Verb}:
 
\begin{tikzpicture}
	\begin{pgfonlayer}{nodelayer}
		\node [style=thick] (0) at (2.75, 3) {};
		\node [style=thick] (1) at (2, 2) {};
		\node [style=thick] (2) at (2.75, 2) {};
		\node [style=thick] (3) at (3.5, 2) {};
		\node [draw, circle, minimum size=0.2 cm, fill=white, style=none] (4) at (2.75, 1.25) {};
		\node [draw, circle, minimum size=0.2 cm, fill=white, style=none] (5) at (2, 0.5) {};
		\node [draw, circle, minimum size=0.2 cm, fill=white, style=none] (6) at (3.5, 0.5) {};
		\node [style=thick] (7) at (1.25, -0.25) {};
		\node [style=thick] (8) at (2.5, -0.25) {};
		\node [style=thick] (9) at (3, -0.25) {};
		\node [style=thick] (10) at (4.25, -0.25) {};
		\node  [draw, circle, minimum size=0.2 cm, fill=white, style=none] (11) at (2.75, -0.75) {};
		\node [style=thick] (12) at (2.75, -1.5) {};
	\end{pgfonlayer}
	\begin{pgfonlayer}{edgelayer}
		\draw [thick, looseness=0.00] (2.center) to (4.center);
		\draw [thick, looseness=0.00] (1.center) to (0.center);
		\draw [thick, looseness=0.00] (0.center) to (3.center);
		\draw [thick, looseness=0.00] (1.center) to (3.center);
		\draw [style=thick, bend left=90, looseness=1.75] (5.center) to (6.center);
		\draw [style=thick, bend left=90, looseness=1.75] (7.center) to (8.center);
		\draw [style=thick, bend left=90, looseness=1.75] (9.center) to (10.center);
		\draw [style=thick, bend left=270, looseness=2.50] (8.center) to (9.center);
		\draw [style=thick] (11.center) to (12.center);
	\end{pgfonlayer}
\end{tikzpicture}}

 \hspace{2.5cm}
 \textbf{Sentence}:
  
\begin{tikzpicture}
	\begin{pgfonlayer}{nodelayer}
		\node [style=none] (0) at (0, 3) {};
		\node [style=none] (1) at (2.75, 3) {};
		\node [style=none] (2) at (5.5, 3) {};
		\node [style=none] (3) at (-0.75, 2) {};
		\node [style=none] (4) at (0, 2) {};
		\node [style=none] (5) at (0.75, 2) {};
		\node [style=none] (6) at (2, 2) {};
		\node [style=none] (7) at (2.75, 2) {};
		\node [style=none] (8) at (3.5, 2) {};
		\node [style=none] (9) at (4.75, 2) {};
		\node [style=none] (10) at (5.5, 2) {};
		\node [style=none] (11) at (6.25, 2) {};
		\node [draw, circle, minimum size=0.2 cm, fill=white, style=none] (12) at (2.75, 1.25) {};
		\node [draw, circle, minimum size=0.2 cm, fill=white, style=none] (13) at (2, 0.5) {};
		\node [draw, circle, minimum size=0.2 cm, fill=white, style=none] (14) at (3.5, 0.5) {};
		\node [style=none] (15) at (1.25, -0.25) {};
		\node [style=none] (16) at (2.5, -0.25) {};
		\node [style=none] (17) at (3, -0.25) {};
		\node [style=none] (18) at (4.25, -0.25) {};
		\node [style=none] (19) at (0, -0.25) {};
		\node [style=none] (20) at (5.5, -0.25) {};
		\node [draw, circle, minimum size=0.2 cm, fill=white, style=none] (21) at (2.75, -0.75) {};
		\node [style=none] (22) at (2.75, -1.5) {};
	\end{pgfonlayer}
	\begin{pgfonlayer}{edgelayer}
		\draw [thick, looseness=0.00] (9.center) to (2.center);
		\draw [thick, looseness=0.00] (9.center) to (11.center);
		\draw [thick, looseness=0.00] (3.center) to (5.center);
		\draw [thick, looseness=0.00] (7.center) to (12.center);
		\draw [thick, looseness=0.00] (6.center) to (1.center);
		\draw [thick, looseness=0.00] (0.center) to (5.center);
		\draw [thick, looseness=0.00] (1.center) to (8.center);
		\draw [thick, looseness=0.00] (3.center) to (0.center);
		\draw [thick, looseness=0.00] (6.center) to (8.center);
		\draw [thick, looseness=0.00] (2.center) to (11.center);
		\draw [style=thick, bend left=90, looseness=1.75] (13.center) to (14.center);
		\draw [style=thick, bend left=90, looseness=1.75] (15.center) to (16.center);
		\draw [style=thick, bend left=90, looseness=1.75] (17.center) to (18.center);
		\draw [style=thick] (4.center) to (19.center);
		\draw [style=thick] (10.center) to (20.center);
		\draw [style=thick, bend left=270, looseness=2.25] (18.center) to (20.center);
		\draw [style=thick, bend right=90, looseness=2.50] (19.center) to (15.center);
		\draw [style=thick, bend left=270, looseness=2.50] (16.center) to (17.center);
		\draw [style=thick] (21.center) to (22.center);
	\end{pgfonlayer}
\end{tikzpicture}}

\end{center}

\vspace{-0.2cm}
\noindent The  normal form of the diagram of the sentence  and its closed linear algebraic form are as follows:
\vspace{-0.2cm}
\begin{center}
 
\begin{tikzpicture}
	\begin{pgfonlayer}{nodelayer}
		\node [style=none] (0) at (1.75, 3) {};
		\node [style=none] (1) at (3.75, 3) {};
		\node [style=none] (2) at (5.75, 3) {};
		\node [style=none] (3) at (1, 2) {};
		\node [style=none] (4) at (1.75, 2) {};
		\node [style=none] (5) at (2.5, 2) {};
		\node [style=none] (6) at (3, 2) {};
		\node [style=none] (7) at (3.75, 2) {};
		\node [style=none] (8) at (4.5, 2) {};
		\node [style=none] (9) at (5, 2) {};
		\node [style=none] (10) at (5.75, 2) {};
		\node [style=none] (11) at (6.5, 2) {};
		\node [draw, circle, minimum size=0.2 cm, fill=white, style=none] (12) at (3.75, 0.75) {};
		\node [style=none] (13) at (3.75, -0) {};
	\end{pgfonlayer}
	\begin{pgfonlayer}{edgelayer}
		\draw [thick, looseness=0.00] (9.center) to (2.center);
		\draw [thick, looseness=0.00] (9.center) to (11.center);
		\draw [thick, looseness=0.00] (3.center) to (5.center);
		\draw [thick, looseness=0.00] (7.center) to (13.center);
		\draw [thick, bend left=270, looseness=1.00] (4.center) to (10.center);
		\draw [thick, looseness=0.00] (6.center) to (1.center);
		\draw [thick, looseness=0.00] (0.center) to (5.center);
		\draw [thick, looseness=0.00] (1.center) to (8.center);
		\draw [thick, looseness=0.00] (3.center) to (0.center);
		\draw [thick, looseness=0.00] (6.center) to (8.center);
		\draw [thick, looseness=0.00] (2.center) to (11.center);
	\end{pgfonlayer}
\end{tikzpicture}}
 
 \hspace{1cm} = \hspace{1cm} 
 $\ov{sbj} \odot \ov{verb} \odot \ov{obj}$
\end{center}


\paragraph{{\sc \textbf{Kron}}} In the model of \cite{GrefenSadr2}, the tensor of a transitive sentence is calculated as the Kronecker product of the context vector of the verb with itself, so we have $\overline{verb} = \ov{verb} \otimes \ov{verb}$. To encode this, we start from the Kronecker product of the context vector of the verb with itself, apply one $\sigma $ to each one of the vectors and then a $\mu$ to both of them jointly. The result is the following left hand side verb, which when substituted in the equation of Definition(*) results in a normal form (depicted in the right hand side below) very similar to the normal form of the {\sc Multp} model:

\vspace{-0.2cm}
\begin{center}
{\bf Verb}: 
\begin{tikzpicture}
	\begin{pgfonlayer}{nodelayer}
		\node [draw, circle, minimum size=0.2 cm, fill=white, style=none] (0) at (-3.5, 1) {};
		\node [style=none] (1) at (-4.25, 0.25) {};
		\node [style=none] (2) at (-2.75, 0.25) {};
		\node [draw, circle, minimum size=0.2 cm, fill=white, style=none] (3) at (-2, -0.5) {};
		\node[style=none] (4) at (-0.5, 1) {};
		\node [style=none](5) at (0.25, 0.25) {};
		\node [style=none] (6) at (-1.25, 0.25) {};
		\node [style=none] (7) at (-4.25, 1.75) {};
		\node [style=none] (8) at (-2.75, 1.75) {};
		\node [style=none] (9) at (-3.5, 2.75) {};
		\node [style=none] (10) at (-3.5, 1.75) {};
		\node [style=none] (11) at (-0.5, 1.75) {};
		\node [style=none] (12) at (-1.25, 1.75) {};
		\node [draw, circle, minimum size=0.2 cm, fill=white, style=none] (13) at (-0.5, 1) {};
		\node [style=none] (14) at (0.25, 1.75) {};
		\node [style=none] (15) at (-0.5, 2.75) {};
		\node [style=none] (16) at (-2, -1.5) {};
	\end{pgfonlayer}
	\begin{pgfonlayer}{edgelayer}
		\draw [style=thick, bend left=90, looseness=1.75] (1.center) to (2.center);
		\draw [style=thick, bend left=90, looseness=1.75] (6.center) to (5.center);
		\draw [style=thick] (7.center) to (8.center);
		\draw [style=thick] (9.center) to (7.center);
		\draw [style=thick] (9.center) to (8.center);
		\draw [style=thick] (10.center) to (0.center);
		\draw [style=thick] (12.center) to (14.center);
		\draw [style=thick] (15.center) to (12.center);
		\draw [style=thick] (15.center) to (14.center);
		\draw [style=thick] (11.center) to (13.center);
		\draw [style=thick, bend right=90, looseness=1.75] (2.center) to (6.center);
		\draw [style=thick] (3) to (16.center);
	\end{pgfonlayer}
\end{tikzpicture}}

\hspace{2cm}
{\bf Sentence}:

\begin{tikzpicture}
	\begin{pgfonlayer}{nodelayer}
		\node [style=none] (0) at (2, 3) {};
		\node [style=none] (1) at (3.75, 3) {};
		\node [style=none] (2) at (5.5, 3) {};
		\node [style=none] (3) at (7.25, 3) {};
		\node [style=none] (4) at (1.25, 2) {};
		\node [style=none] (5) at (2, 2) {};
		\node [style=none] (6) at (2.75, 2) {};
		\node [style=none] (7) at (3, 2) {};
		\node [style=none] (8) at (3.75, 2) {};
		\node [style=none] (9) at (4.5, 2) {};
		\node [style=none] (10) at (4.75, 2) {};
		\node [style=none] (11) at (5.5, 2) {};
		\node [style=none] (12) at (5.5, 2) {};
		\node [style=none] (13) at (6.25, 2) {};
		\node [style=none] (14) at (6.5, 2) {};
		\node [style=none] (15) at (7.25, 2) {};
		\node [style=none] (16) at (8, 2) {};
		\node [draw, circle, minimum size=0.2 cm, fill=white, style=none] (17) at (4.65, 0.8) {};
		\node [style=none] (18) at (4.65, -0) {};
	\end{pgfonlayer}
	\begin{pgfonlayer}{edgelayer}
		\draw [thick, looseness=0.00] (10.center) to (2.center);
		\draw [thick, bend right=90, looseness=0.75] (5.center) to (15.center);
		\draw [thick, looseness=0.00] (2.center) to (13.center);
		\draw [thick, bend right=90, looseness=2.25] (8.center) to (11.center);
		\draw [thick, looseness=0.00] (3.center) to (16.center);
		\draw [thick, looseness=0.00] (7.center) to (1.center);
		\draw [thick, looseness=0.00] (4.center) to (0.center);
		\draw [thick, looseness=0.00] (7.center) to (9.center);
		\draw [thick, looseness=0.00] (14.center) to (3.center);
		\draw [thick, looseness=0.00] (0.center) to (6.center);
		\draw [thick, looseness=0.00] (17.center) to (18.center);
		\draw [thick, looseness=0.00] (14.center) to (16.center);
		\draw [thick, looseness=0.00] (10.center) to (13.center);
		\draw [thick, looseness=0.00] (4.center) to (6.center);
		\draw [thick, looseness=0.00] (1.center) to (9.center);
	\end{pgfonlayer}
\end{tikzpicture}}

\end{center}
\vspace{-0.2cm}

\noindent 

\noindent Linear algebraically, the above normal form is  equivalent to:

\vspace{-0.2cm}
\begin{equation}
\ov{sbj} \odot \ov{verb} \odot \ov{verb} \odot \ov{obj}
\end{equation}

\section{Experiments}
\label{sec:exp}

The different options presented in Section \ref{sec:frobop} and summarized in Table~\ref{tbl:models}  provide us a  number of models for testing our setting. We train our vectors on the  British National Corpus (BNC), which has about six  million sentences and one million words, classified into a hundred million different lexical tokens. We use the set of its 2000 most frequent lemmas as a basis of our basic distributional vector space $W$. The weights of each vector are set to the ratio of the probability of the context word given the target word to the probability of the context word overall. As our similarity measure we use the cosine distance between the vectors. 

\begin{table}[h]
\small
\begin{center}
\begin{tabular}{ll}
  \hline
   \textbf{Model} & \textbf{Description} \\
  \hline\hline
  {\sc CpSbj} & Copy subject on relational matrices \\
  {\sc CpObj} & Copy object on relational matrices \\
  {\sc MixCpDl} & Diagonalize on relational matrices \\
  {\sc Kron} & Diagonalize on direct matrices \\
  {\sc Multp} & Multiplicative model \\
  \hline
\end{tabular} 
\caption{The models.}
\label{tbl:models}
\end{center}
\normalsize
\end{table}

\subsection{Disambiguation} 
\label{sec:disamb}

We first test our models on a disambiguation task, which is an extension of Sch\"utze's original disambiguation task from words to sentences. A dataset for this task was originally developed in \cite{Lapata} for intransitive sentences, and later extended to transitive sentences in \cite{GrefenSadr1}; we use the latter. The goal is to assess how well a model can discriminate between the different senses of an ambiguous verb, given the context (subject and object) of that verb. The entries of this dataset consist of a target verb, a subject, an object, and a landmark verb used for the comparison. One such entry for example is, ``write, pupil, name, spell''. A good  model should be able to understand that the sentence ``pupil write name'' is closer to the sentence ``pupil spell name'' than, for example, to ``pupil publish name''. On the other hand, given the context ``writer, book'' these results should be reversed. 
The evaluation of this experiment is performed by calculating Spearman's $\rho$, which measures the degree of  correlation of the cosine distance with the judgements of 25 human evaluators, who has been asked to assess the similarity of each pair of sentences using a scale from 1 to 7. As our baseline we use a simple additive model ({\sc Addtv}), where the meaning of a transitive sentence is computed as the addition of the relevant context vectors. 

\paragraph{Results} The results of this experiment are shown in Table \ref{tbl:disres}, indicating that the most successful model for this task is the copy-object model. 
The better performance of this model against the copy-subject approach provides us some insights about the role of subjects and objects in disambiguating our verbs. By copying the dimension associated with the object, the compression of the original sentence matrix, as this was calculated in \cite{GrefenSadr1}, takes place along the dimension of subjects (rows), meaning that the resulting vector will bring much more information from the objects than the subjects (this is also suggested by Equation {\ref{equ:clfrm_obj}). Hence, the fact that this vector performs better than the one of the copy-subject method provides an indication that the object of some ambiguous verbs (which turns out to be the case for our dataset) can be more important for disambiguating that verb than the subject. Intuitively, we can imagine that the crucial factor to disambiguate between the verbs ``write'', ``publish'' and ``spell'' is more the object than the subject: a book or a paper can be both published and written, but a letter or a shopping list can only be written. Similarly, a word can be spelled and written, but a book can only be written. The subject in all these cases is not so important. 

\begin{table}
\small
\begin{center}
\begin{tabular}{l|ccc}
\hline
\textbf{Model} & \textbf{High} & \textbf{Low} & $\rho$ \\
\hline\hline
{\sc Addtv} & 0.90 & 0.90 & 0.050 \\ 
\hline
{\sc Multp} & 0.67 & 0.60 & 0.163 \\
{\sc MixCpDl} & 0.75 & 0.77 & 0.000 \\
{\sc Kron} & 0.31 & 0.21 & 0.168 \\
{\sc CpSbj} & 0.95 & 0.95 & 0.143 \\    
{\sc CpObj} & 0.89 & 0.90 & 0.172 \\
\hline
UpperBound & 4.80 & 2.49 & 0.620 \\
\hline
\end{tabular}
\caption{Disambiguation results. High and Low refer to average scores for high and low landmarks, respectively. UpperBound refers to agreement between annotators.}
\label{tbl:disres}
\end{center}
\normalsize
\end{table}


The copy-object model is followed closely by the  ({\sc Kron})  and the {\sc Multp} models. The similar performance of these two  is not a surprise, given their almost identical nature. Finally, the bad performance of the model ({\sc MixCpDl}) that is obtained  by the application of the uncopying $\mu$ map  conforms to the predictions of the theory, as these were expressed in Section \ref{sec:frobop}.

\subsection{Comparing Transitive and Intransitive Sentences}


In this section we will examine the potential of the above approach in practice, in the context of an experiment aiming to compare transitive and intransitive sentences. In order to do that, we use the dataset of the previous verb disambiguation task (see detailed description in Section \ref{sec:disamb}) to conduct the following simple experiment: We create intransitive versions of all the transitive sentences from target verbs and their high and low landmarks by dropping the object; then, we compare each transitive sentence coming from the target verbs with all the other intransitive sentences, expecting that the highest similarity would come from its own intransitive version, the next higher similarity would come from the intransitive version that uses the corresponding high landmark verb, and so on. To present a concrete example, consider the entry ``write, pupil, name, spell, publish''. Our transitive sentence here is $s_{tr} =$ ``pupil write name''; the intransitive version of this is $s_{in}=$ ``pupil write''. We also create intransitive versions using the high and the low landmarks, getting $s_{hi}=$ ``pupil spell'' and $s_{lo}=$``pupil publish''. If the similarity between two sentences $s_1$ and $s_2$ is given by $sim(s_1,s_2)$, we would expect that:

\[
  sim(s_{tr},s_{it}) > sim(s_{tr},s_{hi}) > sim(s_{tr},s_{lo}) > sim(s_{tr},s_{u})
\]
 
\noindent where $s_{u}$ represents an unrelated intransitive version coming from a target verb different than the one of $s_{tr}$. The results of this experiment are shown in Table \ref{tbl:exp} below, for 100 target verbs.

\begin{table}[h]
\begin{center}
\begin{tabular}{lll}
 \hline
 \textbf{Case} & \textbf{Errors} & \% \\
 \hline\hline
 $sim(s_{tr},s_{hi}) > sim(s_{tr},s_{it})$ & 7 of 93 & 7.5\\
 $sim(s_{tr},s_{lo}) > sim(s_{tr},s_{it})$ & 6 of 93 & 5.6\\
 $sim(s_{tr},s_{u}) > sim(s_{tr},s_{it})$ & 36 of 9900 & 0.4\\
 \hline
\end{tabular}
\caption{Results of the comparison between transitive and intransitive sentences.}
\label{tbl:exp}
\end{center}
\end{table}

The outcome follows indeed our expectations for this task. We see, for example, that the highest error rate comes from cases where the intransitive sentence of the high-landmark verb is closer to a transitive sentence than the intransitive version coming from the sentence itself (first row of the table). Since the meaning of a target verb and the high-landmark verb were specifically selected to be very similar given the context (subject and object), this is naturally the most error-prone category. The seven misclassified cases are presented in Table \ref{tbl:fail01}, where the similarity of the involved intransitive versions is apparent.

\begin{table}[h]
\begin{center}
\begin{tabular}{lll}
 \hline
 $s_{tr}$ & $s_{in}$ & $s_{hi}$ \\
 \hline\hline
 people run round & people run & people move \\
 boy meet girl & boy meet & boy visit \\
 management accept responsibility & management accept & management bear \\
 patient accept treatment & patient accept & patient bear \\
 table draw eye & table draw & table attract \\
 claimant draw benefit & claimant draw & claimant attract \\
 tribunal try crime & tribunal try & tribunal judge \\
 \hline
\end{tabular}
\caption{Errors  in the first category of comparisons.}
\label{tbl:fail01}
\end{center}
\end{table}

The six cases of the second category (where an intransitive sentence from a low-landmark gives higher similarity than the normal intransitive version) are quite similar, since in many cases dropping the object leads to semantically identical expressions. For the transitive sentence ``tribunal try crime'', for example, the low-landmark intransitive version ``tribunal test'' has almost identical meaning with the normal intransitive version ``tribunal try'', so it is easier to be ``mistakenly'' selected by the model as the one closest to the original transitive sentence. 

Finally, the model performs really well for cases when an unrelated intransitive sentence is compared with a transitive one, with only a 0.4\% error rate. Here many of the misclassifications can also be attributed to the increased ambiguity of the involved verbs when the object is absent. For example, the similarity between ``man draw sword'' and ``man draw'' is considered smaller than the similarity of the first sentence with ``man write''. Although this is an obvious error, we should acknowledge that the two intransitive sentences, ``man draw'' and ``man write'', are not so different semantically, so the error was not completely unjustified.

\subsection{Definition Classification}

The ability of reliably comparing the meaning of single words with larger textual fragments, e.g. phrases or even sentences, can be an invaluable tool for many challenging NLP tasks, such as definition classification, sentiment analysis, or even the simple everyday search on the internet. In this task we examine the extend to which our models can correctly match a number of terms (single words) with a number of definitions. Our dataset consists of 112 terms (72 nouns and 40 verbs) extracted from a junior dictionary together with their main definition. For each term we added two more definitions, either by using  entries of WordNet for the term or by simple paraphrase of the main definition using a thesaurus, getting a total of three definitions per term. In all cases a definition for a noun-term is a noun phrase, whereas the definitions for the verb-terms consist of verb phrases. 
A sample of the dataset entries can be found in Table \ref{tbl:dfnsample}. 

%

\begin{table}[h]
\begin{center}
\small
\begin{tabular}{l|lll}
 \hline
 \textbf{Term}  & \textbf{Main definition} & \textbf{Alternative def. 1} & \textbf{Alternative def. 2} \\
 \hline\hline
 blaze & large strong fire & huge potent flame & substantial heat \\
 husband & married man & partner of a woman & male spouse \\
 foal  & young horse & adolescent stallion & juvenile mare \\
 horror & great fear & intense fright & disturbing feeling \\
 \hline
 apologise & say sorry & express regret or sadness & acknowledge shortcoming or failing \\
 embark    & get on a ship & enter boat or vessel & commence trip \\
 vandalize & break things & cause damage & produce destruction \\
 regret    & be sad or sorry & feel remorse & express dissatisfaction \\
 \hline 
\end{tabular}
\caption{Sample of the dataset for the term/definition comparison task (noun-terms in the top part, verb-terms in the bottom part).}
\label{tbl:dfnsample}
\end{center}
\normalsize
\end{table}

We approach this evaluation problem as a classification task, where  terms have the role of  classes. First, we calculate the distance between each definition and every term in the dataset. The definition is ``classified'' to the term that gives the higher similarity. Due to the nature the dataset, this task did not require  human annotation and we evaluate the results by calculating separate F1-scores for each term, and getting their average as an overall score for the whole model. The results are presented in Table \ref{tbl:dfnresults}.

\begin{table}[b]
\begin{center}
\small
\begin{tabular}{l|ccc||ccc}
 \hline
     & \multicolumn{3}{c||}{\textbf{Nouns}} & \multicolumn{3}{|c}{\textbf{Verbs}} \\
   \textbf{Model} & P & R & F1 & P & R & F1 \\ 
   \hline\hline
 {\sc Addtv} & 0.21 & 0.17 & 0.16 & 0.28 & 0.25 & 0.23 \\
 \hline
 {\sc Multp} & 0.21 & 0.22 & 0.19 & 0.31 & 0.30 & 0.26 \\
 {\sc Reltn} & 0.22 & 0.24 & 0.21 & 0.32 & 0.28 & 0.27 \\
 \hline 
\end{tabular}
\caption{Results of the term/definition comparison task.}
\label{tbl:dfnresults}
\end{center}
\normalsize
\end{table}

\paragraph{Results} Since this experiment includes verb phrases, where the subject is missing, we construct our verb vectors by summing over all context vectors of objects with which the verb appears in the corpus; that is, we use $\overline{verb} = \sum_i \ov{obj_i}$. This is referred to as the relational model ({\sc Reltn}), and is compared with the multiplicative model. Additive model serves again as our baseline. We evaluate separately the performance on the noun terms and the performance on the verb terms, since a mixing of the two sets would be inconsistent.

The relational model delivers again the best performance, although the difference from the multiplicative model is small. All models perform better on the verb terms than the noun part of the dataset, yet in general F-scores tend to be low. This is natural, since the challenge that this task poses to a machine is great, and F-score considers anything but the perfect result (every definition assigned to the correct term) as unacceptable. 

An error analysis shows that for the noun-term set the relational model returns the correct main definition in 25 of the 72 cases, whereas in 47 cases (65\%) the correct definition is in the top-five list for that term (Table \ref{tbl:dfnresults01}). The multiplicative model performs similarly, and better for the verb-term set. For this experiment we also calculated Mean Reciprocal Ranks values, which again were very close for the two models. Furthermore, some of the ``misclassified'' cases can also be considered as somehow ``correct''. For example, the definition we originally assigned to the term `jacket' was `short coat'; however, the system ``preferred'' the definition `waterproof cover', which is also correct.  Some interesting other  cases are presented in Table \ref{tbl:dfntbl}. 


 
\begin{table}
\begin{center}
\small
\begin{tabular}{l||l|cc||cc}
 \hline
  & & \multicolumn{2}{c||}{\textsc{Multp}} & \multicolumn{2}{|c}{\textsc{Reltn}} \\
  & \textbf{Rank} & Count & \% & Count & \% \\ 
 \hline\hline
  \multirow{4}{*}{\textbf{Nouns}}
  & 1     & 26 & 36.1 & 25 & 34.7 \\
  & 2-5   & 20 & 27.8 & 22 & 30.6 \\
  & 6-10  & 11 & 15.3 & 5  & 6.9  \\
  & 11-72 & 15 & 20.8 & 20 & 27.8 \\
 \hline  
 \multirow{4}{*}{\textbf{Verbs}}
  & 1     & 15 & 37.5 & 8  & 20.0 \\
  & 2-5   & 10 & 25.0 & 13 & 32.5 \\
  & 6-10  & 6  & 15.0 & 4  & 10.0 \\
  & 11-40 & 9  & 22.5 & 15 & 37.5 \\
 \hline  
\end{tabular}
\caption{Results of the term/definition comparison task based on the rank of the main definition.}
\label{tbl:dfnresults01}
\end{center}
\normalsize
\end{table}

\begin{table}
\begin{center}
\small
\begin{tabular}{lll}
 \hline
 \textbf{Term}  & \textbf{Original definition} & \textbf{Assigned definition} \\
 \hline\hline
 rod    & fishing stick & round handle \\
 jacket & short coat & waterproof cover \\
 mud    & wet soil   & wet ground \\
 labyrinth & complicated maze & burial chamber\\
 \hline
\end{tabular}
\caption{A sample of ambiguous cases where the model assigned a different definition than the original.}
\label{tbl:dfntbl}
\end{center}
\normalsize
\end{table}

\section{Conclusion and Future Work}
\label{sec:futwork}

In summary, after a brief review of the  definitions of compact closed categories and monoidal functors, we ramified their applications to natural language syntax and semantics. We recasted the categorical setting of \cite{Coeckeetal}, which was based on the product of the category of a pregroup type-logic with the category of finite dimensional vector spaces, in terms of a monoidal functor from the former to the latter. This passage is similar to the vector space representation of category of manifolds in a topological quantum field theory.  We showed how the operations of Frobenius algebras over vector spaces provide a concrete instantiation of this setting and used their pictorial calculus to simplify the multi-linear algebraic computations. This instantiation resulted in  meanings of all sentences living in a basic vector space $W$, hence we became able to compare their meanings with one another and also with meanings of single words and phrases. We developed  experiments based on this instantiation and evaluated the predictions of our model in a number of sentence and phrase synonymy tasks. 

We conclude that the concrete setting of this paper provides a robust and scalable base for an implementation of the framework of \cite{Coeckeetal}, ready for further experimentation and applications. It overcomes the shortcomings of our first implementation \cite{GrefenSadr1}, whose main problem was that the vector space representation of the atomic type $s$ was taken to be the tensor space $W \otimes W$ (for a transitive sentence), hence the logical and the concrete types did not match. As a consequence, sentences with nested structures such as `Mary saw John reading a book' could not be assigned a meaning; furthermore, meanings of phrases and sentences with different grammatical structure lived in different vector spaces, so a direct comparison of them was not possible.

An experimental future direction is a higher order evaluation of the definition classification task using an unambiguous vector space along the lines of \cite{Schutze}, where each word is associated with one or more \emph{sense} vectors. A model like this will avoid encoding different meanings of words in one vector, and will help us separate the two distinct tasks of composition and disambiguation that currently are interwoven in a single step.

From the theoretical perspective, one direction is to start from type-logical grammars that are more expressive than pregroups. In recent work \cite{Sadrzadeh} we have shown how the functorial passage to {\bf FVect} can be extended from a pregroup algebra to the Lambek Calculus \cite{Lambek0}, which has a monoidal (rather than compact) structure. It remains to show how this passage can be extended to more expressive versions of Lambek calculi, such as Lambek-Grishin algebras \cite{Moortgat}, calculi of discontinuity \cite{Morrill}, or abstract categorial grammars \cite{Groote}. More specifically, one wants to investigate if the new operations and axioms of these  extensions are preserved by the vector space semantic functor. 

Another  venue  to explore is meanings of logical words, where any context-dependant method fails to succeed. The abstract operations of Frobenius algebras have been used to model a limited class of logical operations on vector spaces \cite{CoeckeVic}, hence these might turn out to be promising in this aspect too.

\end{document}